\documentclass[journal]{IEEEtran}

\usepackage[utf8]{inputenc}
\usepackage{times}
\usepackage{epsfig}
\usepackage{epstopdf}
\usepackage{graphicx}
\usepackage{amsmath}
\usepackage{amssymb}
\usepackage{color,soul}
\usepackage{colortbl}
\usepackage{algorithm}
\usepackage{algorithmic}
\usepackage{multirow}

\usepackage{comment}
\usepackage[misc]{ifsym}

\usepackage{tikz}

\newcommand{\specialcell}[2][c]{%
  \begin{tabular}[#1]{@{}c@{}}#2\end{tabular}}
\newcommand{\specialcelll}[2][l]{%
  \begin{tabular}[#1]{@{}l@{}}#2\end{tabular}}  

\graphicspath{{./figures/}}

\definecolor{dgray}{gray}{0.9}
\definecolor{lgray}{gray}{0.95}

\begin{document}

\title{Performance Evaluation Methodology for Long-Term Visual Object Tracking}

\author{Alan Lukežič$^{1*}$, Luka Čehovin Zajc$^{1*}$, Tomáš Vojíř$^2$, Jiří Matas$^2$ and~Matej Kristan$^1$ \\
$^1$ Faculty of Computer and Information Science, University of Ljubljana, Slovenia \\
$^2$ Faculty of Electrical Engineering, Czech Technical University in Prague, Czech Republic \\
{\tt\small \{alan.lukezic, luka.cehovin\}@fri.uni-lj.si}
\IEEEcompsocitemizethanks{\IEEEcompsocthanksitem
$^{*}$~The authors contributed equally.
}
}

\markboth{Submitted to a journal on June 2018}%
{Shell \MakeLowercase{\textit{et al.}}: Bare Demo of IEEEtran.cls for IEEE Journals}

\maketitle

\begin{abstract}
A long-term visual object tracking performance evaluation methodology and a benchmark are proposed.
Performance measures are designed by following a long-term tracking definition to maximize the analysis probing strength. The new measures outperform existing ones in interpretation potential and in better distinguishing between different tracking behaviors. 
We show that these measures generalize the short-term performance measures, thus linking the two tracking problems. 
Furthermore, the new measures are highly robust to temporal annotation sparsity and allow annotation of sequences hundreds of times longer than in the current datasets without increasing manual annotation labor. 
A new challenging dataset of carefully selected sequences with many target disappearances is proposed. 
A new tracking taxonomy is proposed to position trackers on the short-term/long-term spectrum. The benchmark contains an extensive evaluation of the largest number of long-term tackers and comparison to state-of-the-art short-term trackers. 
We analyze the influence of tracking architecture implementations to long-term performance and explore various re-detection strategies as well as influence of visual model update strategies to long-term tracking drift. 
The methodology is integrated in the VOT toolkit to automate experimental analysis and benchmarking and to facilitate future development of long-term trackers.
\end{abstract}


\IEEEpeerreviewmaketitle

\section{Introduction}
%
%
%
%
\IEEEPARstart{T}{he} field of visual object tracking has significantly advanced over the last decade. The progress has been helped by the emergence of standard datasets, performance evaluation protocols~\cite{otb_cvpr2013,alov_pami2014,templecolor_tip2015,kristan_vot_tpami2016,MOTChallenge2015} and tracking challenges~\cite{kristan_vot2017,MOTChallenge2015}.

Popular single-target tracking benchmarks~\cite{otb_pami2015,alov_pami2014,templecolor_tip2015,kristan_vot_tpami2016} focus on short-term trackers. 
The introduction of gradually more demanding benchmarks lead to the development of short-term trackers that cope well with significant appearance and motion changes and are robust to short-term occlusions. 
Several recent publications~\cite{uav_benchmark_eccv2016,moudgil2017long,tao2017tracking} show that short-term trackers fare poorly on very long sequences since localization errors and updates gradually deteriorate their visual model, leading to drift and failure. 
Typically, short-term trackers assume that the target is always in the field of view (this is reflected in the standard dataset). When this is not the case, the short-term tracker fails, forever. 

\begin{figure*}[!t]
\begin{center}
	\includegraphics[width=\linewidth]{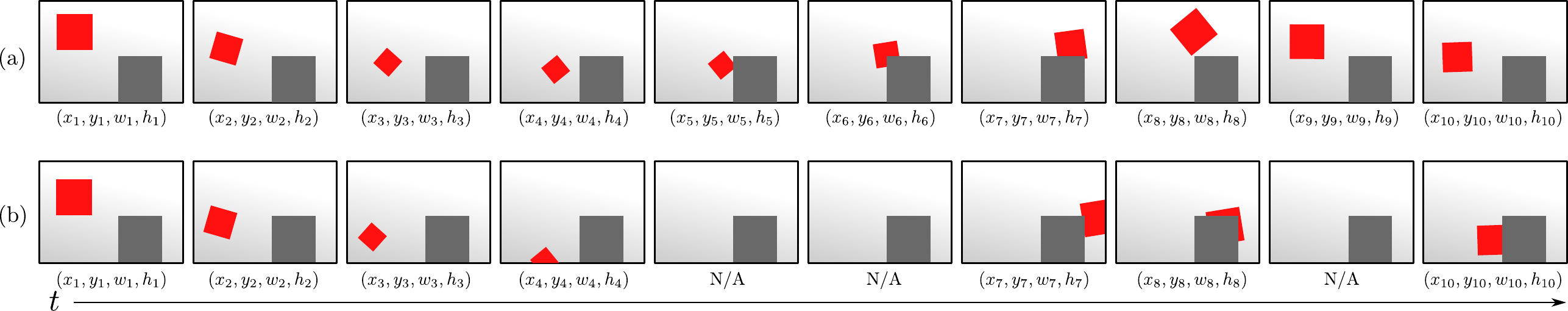}
\end{center}
   \caption{Differences between short-term and long-term tracking. (a) In short-term tracking, the target, a red box, may move and change appearance, but  it is always at least partially visible. (b) In long-term tracking, the box may disappear from the view or be fully occluded by other objects for long periods of time. Within these periods, the state of the object is not defined and should not be reported by the tracker.} 
\label{fig:long-term-tracking} 
\end{figure*}

Long-term trackers are designed for scenarios where the target may disappear from the field of view, may be fully occluded for long periods of time and where cuts, i.e. unpredictable abrupt changes of target pose and appearance, may occur. 
A long-term tracker thus requires to have the ability to report that the target is not present, e.g. by providing a confidence score of the estimated pose, which may be binary or continuous, with low confidence suggesting the target is absent. 

A crucial difference to short-term tracking is thus the re-detection capability, i.e. the ability to localize the target when no information about current poses is available (Figure~\ref{fig:long-term-tracking}). This requires fundamentally different search strategies and  visual model adaptation mechanisms. 
These long-term aspects have been explored far less than the short-term counterparts due to lack of benchmarks and performance measures probing long-term capabilities. This is the focus of our work.

Apart from coping with long sequences, long-term tracking primarily refers to the sequence properties (number of target disappearances, etc.) and the type of tracking output expected. 
We start by defining the notion of the pure long-term tracker and contrast it with  pure short-term tracking. 
We then argue there is a spectrum of tracker designs on the  short-term/long-term axis. 
Based on the long-term definition we propose new performance measures, evaluation protocol and the dataset, all carefully designed to expose the long-term tracking properties. We experimentally show the proposed performance measures produce well-interpretable results. We also show significant robustness to the annotation sparsity. 

Using the proposed evaluation tools, we provide an in-depth analysis of the largest number of long-term trackers to date. 
The analysis includes a new re-detection experiment that exposes crucial long-term tracking capabilities. The tracker performance is analyzed with respect to sequence attributes, the target disappearance rate as well as tracking speed variation. 
Analysis of long-term tracking architectures is provided as well. We test the overall performance of the architectures and analyze re-detection strategies and influence of the model update strategies on the long-term tracking drift. 
The preliminary version of this  work was published in~\cite{lukezic_ltb_arxiv}. We make the following contributions:
\begin{itemize}
	\item A new short-term/long-term tracking taxonomy for fine categorization of long-term trackers. 
	\item A new long-term tracking performance evaluation methodology which introduces novel performance measures {\em tracking} Precision, Recall and F-score. The tracking F-score is a generalization of popular short-term tracking measure.
	To the best of our knowledge, these are the first measures that principally reflect detection as well as localization accuracy in a long-term tracking domain. Comparison with exiting measures shows significant advantages in their expressive power.
    \item A new dataset is constructed  of carefully selected sequences with a large number of target disappearances per sequence to emphasize long-term tracking properties. Sequences are annotated with nine visual attributes which enable in-depth analysis of trackers.
    \item A detailed analysis of the largest number of long-term trackers covering several aspects of long-term properties, sequence attributes, target disappearance rate and speed.
    \item A detailed analysis of long-term tracker architectures from perspective of re-detection and drift-prevention approaches.
\end{itemize}

All trackers, performance measures and evaluation protocol have been integrated into the VOT toolkit~\cite{kristan_vot_tpami2016}, to automate experimental analysis and benchmarking and facilitate development of long-term trackers. 
The dataset, all the trackers as well as the changes to the toolkit will be made publicly available.

The remainder of the paper is structured as follows. 
Section~\ref{sec:related_work} overviews the most closely related work. 
The short-term/long-term taxonomy is presented in Section~\ref{sec:long-term}, 
Section~\ref{sec:methodology} presents the new performance measures and the dataset is presented in Section~\ref{sec:dataset}. 
Section~\ref{sec:measures-analysis} contains analysis of performance measures, 
the new long-term tracking benchmark is presented in Section~\ref{sec:lt-benchmark} 
and Section~\ref{sec:architecture_evaluation} evaluates various tracking architecture designs. 
Conclusions are drawn in Section~\ref{sec:conclusion}.

\section{Related work}  \label{sec:related_work}

Performance evaluation in single-object tracking has primarily focused on short-term trackers~\cite{otb_pami2015,kristan_vot_tpami2016,templecolor_tip2015,alov_pami2014}. 
The currently widely-used methodologies originate from three benchmarks, 
OTB~\cite{otb_cvpr2013,otb_pami2015}, VOT~\cite{kristan_vot2013,kristan_vot_tpami2016} and ALOV~\cite{alov_pami2014} which primarily differ in the dataset construction, performance measures and evaluation protocols.

Benchmarks like \cite{otb_pami2015}, \cite{alov_pami2014}, \cite{templecolor_tip2015} propose large datasets, reasoning that quantity reduces the variance in performance estimation. 
On the other hand, the longest-running benchmark~\cite{kristan_vot_tpami2016} argues that quantity does not necessarily mean quality and promotes moderate-sized datasets with carefully chosen diverse sequences for fast and informative evaluation. Several works have focused on specific tracking setups. 
Mueller et al.~\cite{uav_benchmark_eccv2016} proposed the UAV123 dataset for tracking from drones.
Galoogahi et al.~\cite{Galoogahi_2017_ICCV} introduced a high-frame-rate dataset to analyze trade-offs between tracker speed and robustness. Čehovin et al.~\cite{cehovin_iccv2017} proposed a dataset with an active camera view control using omni directional videos for accurate tracking analysis as a function camera motion attributes. 
The target never leaves the field of view in these datasets, making them unsuitable for long-term tracking properties evaluation. 

Many performance measures have been explored to evaluate and rank single-target trackers~\cite{cehovin_tip2016}. All dominant short-term performance measures~\cite{otb_pami2015,alov_pami2014,kristan_vot_tpami2016} are based on the overlap (intersection over union) between the ground truth bounding boxes and tracker predictions, but significantly differ in its use. 
ALOV \cite{alov_pami2014} uses the F-measure computed at overlap threshold of 0.5. OTB~\cite{otb_pami2015} avoids the threshold by computing the average overlap over the sequences as the primary measure. 
The VOT~\cite{kristan_vot_tpami2016} resets the tracker once the overlap drops to zero, and proposes to measure robustness by the number of times the tracker was reset, the accuracy by average overlap during successful tracking periods and an expected average overlap on a typical short-term sequence. 
These measures do not account for tracker ability to report target absence and are therefore not suitable for long-term tracking.

A large number of performance measures have been proposed for multi-object tracking \cite{MOT16}. 
The two most widely used are MOTA and MOTP~\cite{clear_2006}.
MOTA  is based  on  counting wrong target predictions, defined as:
(i)~target prediction not given (target miss), (ii)~prediction given, but its overlap with the target is too small (false positive prediction) and 
(iii)~a wrong identity assigned to a target prediction (mismatch).
MOTP \cite{clear_2006}  measures the average overlap on frames where target is correctly identified and the overlap is greater than 0.5.
Both measures require setting a threshold that defines
whether the target is successfully located. 
The measure are sensitive to the setting since a small change
of  the threshold may have a large impact on the results \cite{kristan_vot_tpami2016}. 

Another group of measures is based on target trajectories \cite{Bo_cvpr_2006}.
The trajectory of each annotated target in the video is classified into three classes: mostly tracked (MT), partially tracked (PT) and mostly lost (ML). 
A trajectory is mostly tracked if it is correctly tracked in at least 80\% of frames where the target is visible. In mostly lost trajectories the target is tracked in less than 20\% of frames.
All other trajectories are partially tracked.
The ratios of MT, PT and ML to the number of annotated trajectories are reported as measures.
These measures require (ad hoc) thresholds. 
Since the measures are defined on multiple trajectories, applying them to a single trajectory, which is the case in single-target tracking, translates them to success rate \cite{otb_cvpr2013} calculated at specific thresholds.

A few papers have recently proposed datasets focusing on long-term performance evaluation. 
Tao et al.~\cite{tao2017tracking} created artificial long sequences by repeatedly playing shorter sequences forward and backward. 
Such a dataset exposes the problem of gradual drift in short-term trackers, but does not fully expose the long-term abilities since the target never leaves the field of view. 
Mueller et al.~\cite{uav_benchmark_eccv2016} proposed UAV20L dataset of twenty long sequences with target frequently exiting and re-entering the scene, but used it to evaluate mostly short-term trackers. 
A dataset with many cases of fully occluded and absent target has been recently proposed by Moudgil and Gandhi~\cite{moudgil2017long}. 
Unfortunately, the large number of target disappearances was obtained by significantly increasing the sequence length, which significantly increases the storage requirements. To cope with this, a very high video compression is applied, thus sacrificing the image quality. 
   
In the absence of a clear long-term tracking definition, much less attention has been paid to long-term performance measures. 
The UAV20L~\cite{uav_benchmark_eccv2016} and~\cite{moudgil2017long} apply the average overlap measure~\cite{otb_pami2015}, a short-term criterion that does not account for situation when the tracker reports target absence and favors the trackers that report target positions for every frame. 
Tao et al.~\cite{tao2017tracking} adapted this measure by assigning  overlap  of $1$ when the tracker correctly predicts the target absence.  This value is not comparable with tracker accuracy when the target is visible which skews the overlap-based measure. 
Furthermore, reducing the actual tracking accuracy and failure detection to a single overlap score significantly limits the insight it brings.

Long-term tracker analysis requires including sequences, which are much longer than those encountered in short-term tracking evaluation. Target annotation in each frame thus significantly increases the amount of manual labor compared to short-term benchmarks. 
Recently, Mueller et al.~\cite{muller_arxiv2018} considered semi-automatic annotation of short-term sequences used for training localization CNNs. 
They annotate a single frame per-second and interpolate between them by a discriminative correlation filter. 
Given a typical sequence frame-rate, this means they manually annotate only every 25th frame. 
The amount of annotation is reduced and the quality is acceptable for training purposes, but it is not clear whether the interpolation adds bias if such an approach is used for performance evaluation. 

Valmadre et al.~\cite{Valmadre_2018_ECCV} propose to completely avoid interpolation and consider only one frame per-second. 
They argue that sparse annotation is acceptable for long-term tracker evaluation on long sequences. Their experiment on a short-term dataset OTB100~\cite{otb_pami2015} shows that evaluating at every $\sim25$th frame keeps the variance of their tracking performance measure within reasonable bounds. 
Increasing the annotation skipping length increases the variance, which could be addressed by increasing the number of sequences.

\section{The Short-term/Long-term tracking spectrum}  \label{sec:long-term}

A long-term tracker is required to handle target disappearance and reappearance (Figure~\ref{fig:long-term-tracking}). 
Relatively few published trackers fully address the long-term requirements, and  some short-term trackers address them partially. 
We argue that trackers should not be simply classified as short-term or long-term, but they rather cover an entire short-term--long-term \textit{spectrum}. The following taxonomy is used in our experimental section for accurate performance analysis.

\begin{enumerate}
\item  {\bf Short-term tracker} ($\mathrm{ST}_0$). The target position is reported for each frame. The tracker does not implement target re-detection and does not explicitly detect occlusion. Such trackers are likely to fail on the first occlusion as their representation is affected by any occluder. 
\item {\bf Short-term tracker with conservative updating} ($\mathrm{ST}_1$). The target position is reported for each frame. Target re-detection is not implemented, but tracking robustness is increased by selectively updating the visual model depending on a tracking confidence estimation mechanism.
\item {\bf Pseudo long-term tracker} ($\mathrm{LT}_0$). The target position is reported only if the tracker believes the target is visible. The tracker does not implement explicit target re-detection but uses an internal mechanism to identify and report tracking failure.
\item {\bf Re-detecting long-term tracker} ($\mathrm{LT}_1$). The target position is  reported only if the tracker believes the target is visible. The tracker detects tracking failure and implements explicit target re-detection. 
\end{enumerate}

The $\mathrm{ST}_0$ and $\mathrm{ST}_1$ trackers are what is commonly considered a short-term tracker. 
Typical representatives from $\mathrm{ST}_0$ are 
KCF \cite{henriques2015tracking}, DSST \cite{danelljan_dsst_pami}, SRDCF \cite{srdcf_iccv2015}, CSRDCF~\cite{Lukezic_CVPR_2017}, BACF \cite{BACF_ICCV2017} and CREST \cite{crest_ICCV17}, which apply a constant visual model update. Typical examples of $\mathrm{ST}_1$ are NCC \cite{kristan_vot2013}, SiamFC \cite{siamfc_eccv16} and the current state-of-the-art short-term trackers MDNet \cite{mdnet_cvpr2016} and ECO \cite{danelljan_iccv2015_convolutional}. All these trackers apply conservative updating mechanisms, which makes them $\mathrm{ST}_1$ level.
Many short-term trackers can be trivially converted into pseudo long-term trackers ($\mathrm{LT}_0$) by using their visual model similarity scores at the reported target position. 
While straightforward, this offers means to evaluate short-term trackers in the long-term context.

The level $\mathrm{LT}_1$ trackers are the most sophisticated long-term trackers, in that they cover all long-term requirements.
These trackers typically combine two components, a short-term tracker and a detector, and implement an algorithm for their interaction. 
The $\mathrm{LT}_1$ trackers originate from two main paradigms introduced by TLD \cite{kalal_pami} and Alien \cite{Pernici2013}, 
with modern examples CMT \cite{CMT_CVPR2015}, Matrioska \cite{Maresca2013}, HMMTxD \cite{Vojir_CVIU2016}, MUSTER \cite{muster_cvpr2015}, LCT \cite{LCT_CVPR2015}, PTAV \cite{ptav_iccv2017}, and FCLT \cite{fclt_arxiv}. 

Interestingly, two recently published trackers LCT \cite{LCT_CVPR2015} and PTAV \cite{ptav_iccv2017}, that perform well in short-term evaluation benchmarks OTB50 \cite{otb_cvpr2013} and OTB100 \cite{otb_pami2015}, 
are presented as long-term trackers, but experiments in Section~\ref{sec:trackers} show they are in the LT$_0$ class.

\section{Long-term tracking performance measures}  \label{sec:methodology}

A long-term tracking performance measure should reflect the localization accuracy, but unlike short-term measures, it should also capture the accuracy of target detection capabilities (target absence prediction and target re-detection). The latter is not addressed by the standard short-term tracking measures.
In detection literature~\cite{everingham2010pascal}, \textit{precision} and \textit{recall} measures evaluate the detector by considering the amount of predicted bounding boxes whose overlap with the ground truth bounding boxes exceeds a pre-defined threshold. However, threshold-dependent overlap measures
do not fully reflect the tracking accuracy, and should be avoided~\cite{kristan_vot_tpami2016,cehovin_tip2016}. In the following we provide a new formulation of \textit{precision} and \textit{recall} measures which are tailored for tracking domain and avoid the deficiencies of their counterparts from the detection literature. The new measures are rigorously compared to the existing ones in Section~\ref{sec:measures-comparison}.


Let $G_t$ be the ground truth target pose, let $A_t(\tau_\theta)$ be the pose predicted by the tracker, $\theta_t$ the prediction certainty score at time-step $t$ and $\tau_\theta$ be a classification threshold. If the target is absent, the ground truth is an empty set, i.e., $G_t=\emptyset$. 
Similarly, if the tracker did not predict the target or the prediction certainty score is below a classification threshold i.e., $\theta_t < \tau_\theta$,
the output is $A_t(\tau_\theta)=\emptyset$. 
The agreement between the ground truth and prediction is specified by their intersection over union $\Omega(A_t(\tau_\theta), G_t)$\footnote{The output of $\Omega(\cdot, \cdot)$ is 0 if any of the two regions is $\emptyset$.}. In the detection literature, the prediction matches the ground truth if the overlap $\Omega(A_t(\tau_\theta), G_t)$ exceeds a  threshold $\tau_\Omega$. Given the two thresholds $(\tau_\theta, \tau_\Omega)$, the precision $Pr$ and recall $Re$ are defined as
\begin{eqnarray} \label{eq:pr_re_general}
Pr(\tau_\theta, \tau_\Omega) = | \{ t : \Omega(A_t(\tau_\theta), G_t) \geq \tau_\Omega \} | / N_p, \\
Re(\tau_\theta, \tau_\Omega) = | \{ t : \Omega(A_t(\tau_\theta), G_t) \geq \tau_\Omega \} | / N_g,
\end{eqnarray}
where $|\cdot|$ is the cardinality, $N_g$ is the number of frames with $G_t\neq\emptyset$ and $N_p$ is the number of frames with existing prediction, i.e. $A_t(\tau_\theta) \neq \emptyset$.
Note that $N_g$ is defined by ground truth and is constant for a selected sequence, while $N_p$ is a function of the target prediction certainty threshold $\tau_{\theta}$.

In detection literature, the overlap threshold is set to $0.5$ or higher, while recent work~\cite{kristan_vot_tpami2016} has demonstrated that such threshold is over-restrictive and does not clearly indicate a tracking failure in practice. 
A popular short-term performance measure~\cite{otb_cvpr2013}, for example, addresses this by averaging performance over various thresholds, which was shown in~\cite{cehovin_tip2016} to be equal to the average overlap. Using the same approach, we reduce the precision and recall to a single threshold by integrating over $\tau_\Omega$, i.e., 

\begin{eqnarray} \label{eq:pr}
Pr(\tau_\theta) &=& \int_0^1 Pr(\tau_\theta, \tau_\Omega) d\tau_\Omega \\ 
\nonumber &=& \frac{1}{N_p} \sum_{t \in \{ t : A_t(\tau_\theta) \neq \emptyset \}  } \Omega(A_t(\tau_\theta), G_t),
\end{eqnarray}
\begin{eqnarray} \label{eq:re}
Re(\tau_\theta) &=& \int_0^1 Re(\tau_\theta, \tau_\Omega) d\tau_\Omega \\ 
\nonumber &=& \frac{1}{N_g} \sum_{t \in \{ t : G_t \neq \emptyset \}  } \Omega(A_t(\tau_\theta), G_t).
\end{eqnarray}

We call $Pr(\tau_\theta)$ {\em tracking} precision and $Re(\tau_\theta)$ {\em tracking} recall to distinguish them from their detection counterparts. 
Detection-like precision/recall plots can be drawn to analyze the tracking as well as detection capabilities of a long-term tracker (Figure~\ref{fig:average_f_pr_re}). 
Similarly, a standard trade-off between the precision and recall can be computed in form of a {\em tracking} F-measure~\cite{everingham2010pascal}
\begin{eqnarray} \label{eq:f_measure}
F(\tau_\theta) = 2 Pr(\tau_\theta) Re(\tau_\theta) / (Pr(\tau_\theta) + Re(\tau_\theta)),
\end{eqnarray}
\noindent and visualized by the F-score plots (Figure~\ref{fig:average_f_pr_re}). Our primary score for ranking long-term trackers is therefore defined as the highest F-score on the F-score plot, i.e., taken at the tracker-specific optimal threshold. 
This avoids manually-set thresholds in the primary performance measure.
Furthermore, it avoids forcing a tracker to internally threshold its target presence uncertainty and more fairly evaluates different trackers at their optimal performance point.
 
Note that the proposed primary measure (\ref{eq:f_measure}) for the long-term trackers is consistent with the established short-term tracking methodology. 
Consider an $\mathrm{ST}_0$ short-term tracking scenario: the target is always (at least partially) visible and the target position is predicted at each frame with equal certainty. 
In this case our F-measure (\ref{eq:f_measure}) reduces to the average overlap, which is a standard measure in short-term tracking~\cite{otb_cvpr2013,kristan_vot_tpami2016}.

\subsection{Performance evaluation protocol}

A tracker is evaluated on a dataset of several sequences by initializing on the first frame of a sequence and run until the end of the sequence without re-sets. 
The precision-recall curve (\ref{eq:pr},~\ref{eq:re}) is calculated on each sequence and averaged into a single plot. 
This guarantees that the result is not dominated by extremely long sequences. The F-measure plot (\ref{eq:f_measure}) is computed from the average precision-recall plot. 
The evaluation protocol along with plot generation was implemented in the VOT~\cite{kristan_vot_tpami2016} toolkit to automate experiments and thus reduce potential human errors.

\section{The long-term dataset (LTB50)}  \label{sec:dataset}

Table~\ref{tab:datasets} quantifies the long-term statistics of the common short-term and existing long-term tracking datasets. 
Target disappearance is missing in the standard short-term datasets except for UAV123 which contains on average less than one full occlusion per sequence. 
This number increases four-fold in UAV20L~\cite{uav_benchmark_eccv2016} long-term dataset. 
The recent TLP~\cite{moudgil2017long} dataset increases the number of target disappearances by an order of magnitude, but at a cost of increasing the dataset size in terms of the number of frames by more than an order of magnitude, 
i.e. target disappearance events are less frequent in TLP~\cite{moudgil2017long} than in UAV20L~\cite{uav_benchmark_eccv2016}, see Table~\ref{tab:datasets}. 
Moreover, the videos are heavily compressed with many artifacts that affect tracking.

\begin{table}[h]
\begin{center}
\caption{Datasets -- comparison of long-term properties: the number of sequences, the total number of frames, the number of target disappearances (DSP), the average length of disappearance interval (ADL), the average number of disappearances in sequence (ADN). The first four datasets are short-term with virtually no target disappearances, the last column shows the properties of the proposed dataset.}
\label{tab:datasets}
\scalebox{.78}{
\begin{tabular}{l   c   c   c   c   c c   c}
\rowcolor{lgray} \hline
{\bf Dataset} & \specialcell{ALOV \\ 300} & \specialcell{OTB \\ 100} & \specialcell{VOT \\ 2017} & \specialcell{UAV \\ 123} & \specialcell{UAV \\ 20L} & \specialcell{TLP} & \specialcell{LTB50 \\(ours)} \\
\rowcolor{dgray}\hline
{\bf \# sequences} & 315 & 100 & 60 & 123 & 20 & 50 & 50 \\
\rowcolor{lgray} {\bf Frames} & 89364 & 58897 & 21356 & 112578 & 58670 & 676431 & 215294 \\
\rowcolor{dgray} {\bf DSP} & 0 & 0 & 0 & 63 & 40 & 316 & 525 \\
\rowcolor{lgray}{\bf ADL} & 0 & 0 & 0 & 42.6 & 60.2 & 64.1 & 52.0 \\
\rowcolor{dgray}{\bf ADN } & 0 & 0 & 0 & 0.5 & 2 & 6.3 & 10.5 \\
\hline
\end{tabular}
}
\end{center}
\end{table}

\begin{figure}[!th]
\begin{center}
	\includegraphics[width=\linewidth]{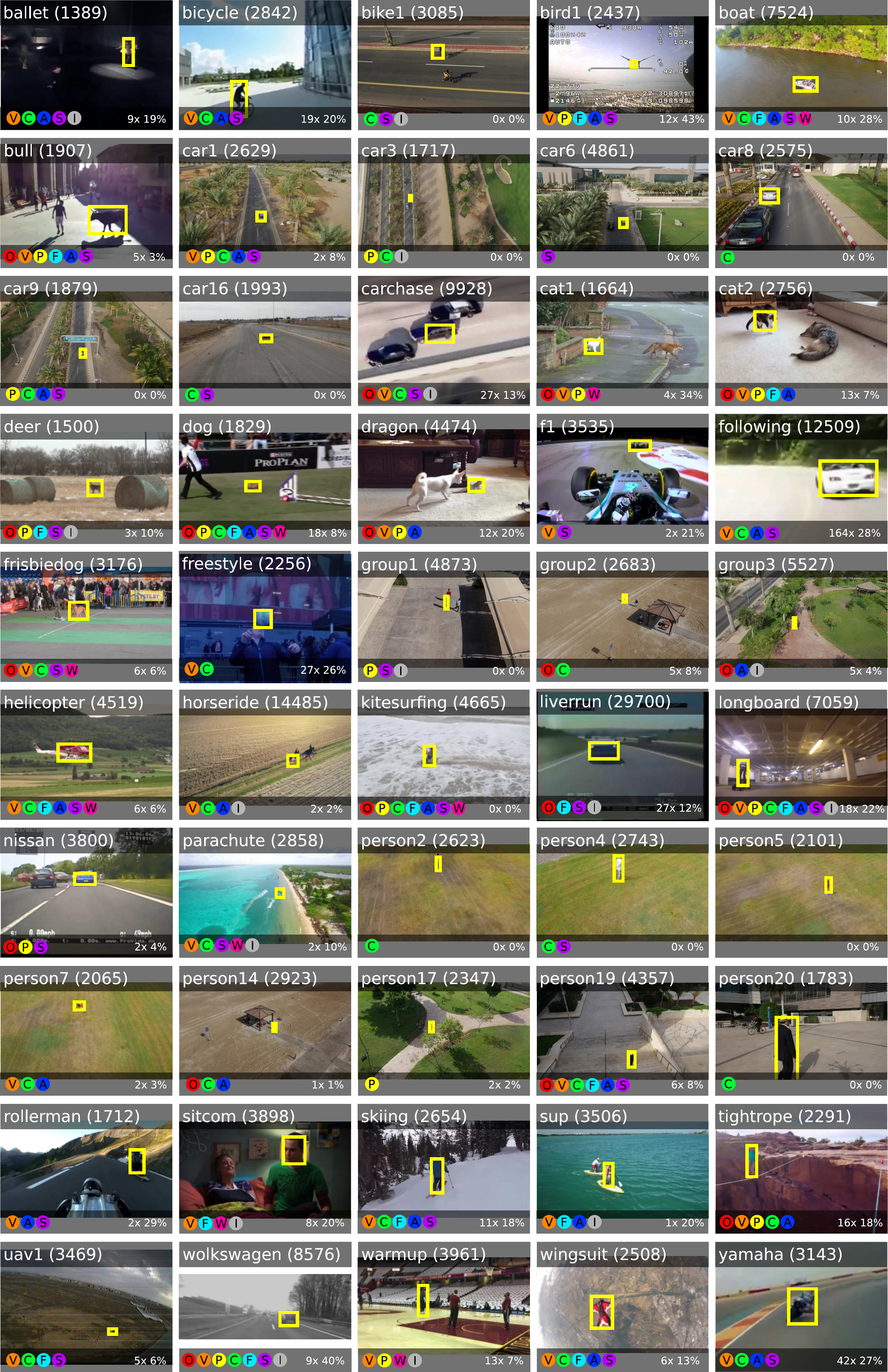}
\end{center}
   \caption{The LTB50 dataset -- a  frame selected from each sequence. Name and length  (top),  number of disappearances and percentage of frames without target (bottom right). Visual attributes (bottom left): (O) Full occlusion, (V) Out-of-view, (P) Partial occlusion, (C) Camera motion, (F) Fast motion, (S) Scale change, (A) Aspect ratio change, (W) Viewpoint change, (I) Similar objects. The dataset is highly diverse in attributes, target types and contains many target disappearances.} 
\label{fig:sequences}
\end{figure}

In the light of the limitations of the existing datasets, we created a new long-term dataset. 
We  followed the VOT~\cite{kristan_vot_tpami2016} dataset construction paradigm 
(recently experimentally validated in~\cite{GOT10k}) 
which states that the datasets should be kept moderately large and manageable, but rich in attributes relevant to the tested tracker class. 
We started by including all sequences from UAV20L since they contain a moderate occurrence  of occlusions and potentially difficult to track small targets. 
Five long sequences with challenging targets were taken from~\cite{kalal_pami}. We collected 19 additional sequences from Youtube. 
The sequences contain larger targets with numerous disappearances. 
To further increase the number of target disappearances per sequence, we have utilized the recently proposed camera view generator from omni-directional dataset AMP \cite{cehovin_iccv2017}. 
Six additional challenging sequences were generated from this dataset by controlling the camera such that the target was repeatedly entering and leaving the field-of-view.

The targets were annotated by axis-aligned bounding-boxes. Each sequence is annotated by nine visual attributes: full occlusion, out-of-view motion, partial occlusion, camera motion, fast motion, scale change, aspect ratio change, viewpoint change and similar objects.
The LTB50 thus contains 50 challenging sequences of diverse objects (persons, car, motorcycles, bicycles, boat, animals, etc.) with the total length of $215294$ frames. 
Sequence resolutions range between $1280 \times 720$ and $290 \times 217$. Each sequence contains on average 10 long-term target disappearances, each lasting on average 52 frames. 
An overview of the dataset is shown in Figure~\ref{fig:sequences}.

\section{Analysis of performance measures}  \label{sec:measures-analysis}

\subsection{Comparison with existing measures}  \label{sec:measures-comparison}

Two threshold-free performance measures were recently used for long-term tracking performance evaluation \cite{uav_benchmark_eccv2016,tao2017tracking,moudgil2017long}. 
The AUC measure is  used in UAV20L \cite{uav_benchmark_eccv2016}. As discussed in Section~\ref{sec:related_work}, this is a primary short-term measure from~\cite{otb_cvpr2013} that computes average overlap between the tracker prediction and ground truth bounding boxes. 
In recent work \cite{tao2017tracking,moudgil2017long} AUC was adapted to account for target absence by assigning overlap of 1 to frames in which the tracker correctly predicts the target absence -- which we denote by AUC$_\mathrm{mod}$. We experimentally compare our long-term performance measures from Section~\ref{sec:methodology} with AUC and AUC$_\mathrm{mod}$ using the approach with theoretical trackers introduced by Čehovin et al.~ \cite{cehovin_tip2016}.

The following four theoretical trackers were run on the LTB50 to expose the differences between the tested performance measures:
\begin{itemize}
	\item $\mathrm{T_{gt,gt}}$: Always reports the correct target position (the ground truth), and reports uncertainty 0 when target is  visible and 1 when target is not visible.
    \item $\mathrm{T_{gt,co}}$: Always reports the correct target position (the ground truth), and reports constant uncertainty in all frames.
    \item $\mathrm{T_{im,co}}$: Reports a bounding box covering entire image in all frames with constant uncertainty, resulting in non-zero overlap in all frames with the target present. The optimal operation point for this tracker is thus to report target always present.
    \item $\mathrm{T_{lost}}$: Reports a 1$\times$1 bounding box in the top-left corner and constant uncertainty in all frames, which is interpreted as if reporting {\it target not visible} in all frames. In contrast to $\mathrm{T_{im,co}}$, which always reports a non-zero overlap, the overlap is always zero for this tracker, thus the optimal operation point is obtained by reporting target always lost.
\end{itemize}

\begin{figure}[!t]
\begin{center}
	\includegraphics[width=\linewidth]{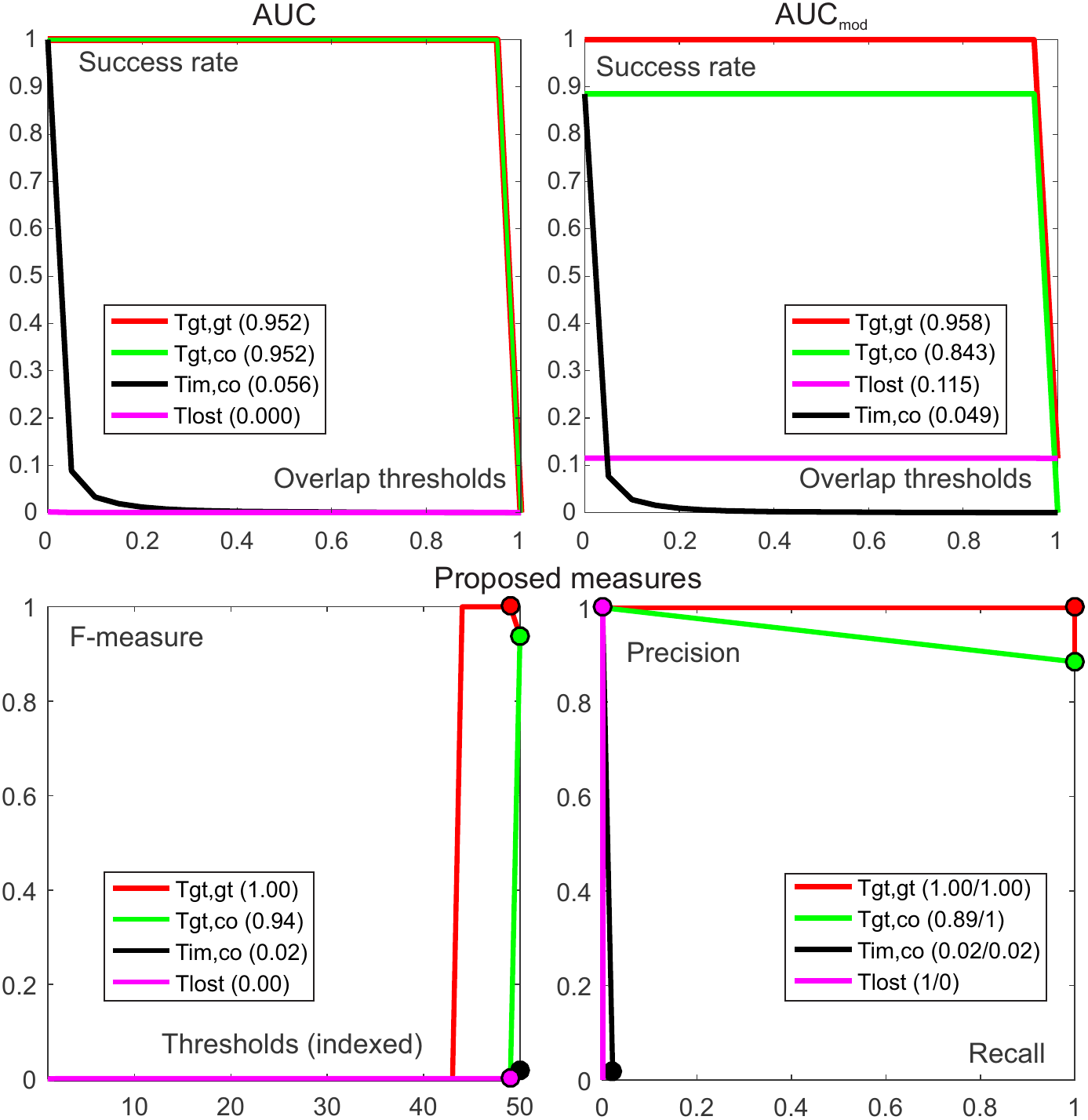}
\end{center}
   \caption{Tracking performance of the four trackers described in Section~\ref{sec:measures-comparison} evaluated by the proposed and existing performance measures. The success rate plots corresponding to AUC and AUC$_\mathrm{mod}$ are shown in the first two plots, while the second two show the tracking F-measure and tracking Precision/Recall graphs. The overall tracker scores are shown in parentheses next to the tracker labels with the two values in the bottom-right plot showing the precision and recall values. } 
\label{fig:theoretical-trackers} 
\end{figure}

Results are summarized in Figure~\ref{fig:theoretical-trackers}. The AUC~\cite{otb_cvpr2013} measure assigns equal scores to $\mathrm{T_{gt,gt}}$ and $\mathrm{T_{gt,co}}$. 
This means it does not distinguish between trackers that can detect target absence and those that cannot. Consequently this measure favors reporting the bounding box in every frame even if the target is not present. 

In contrast, the modified AUC, AUC$_\mathrm{mod}$ from \cite{tao2017tracking} and \cite{moudgil2017long}, does distinguish between $\mathrm{T_{gt,gt}}$ and $\mathrm{T_{gt,co}}$. But this measure assigns a constant overlap 1 to all frames in which the target absence is correctly predicted. 
This is not calibrated by an average overlap when the target is present. 
Furthermore, since the target absence prediction and localization are mixed into a single score, it is unclear whether the high score values are mostly due to accurate prediction of the target position or the ability to correctly report target absence. 
For example, AUC$_\mathrm{mod}$ assigns an average overlap of 11\% to tracker $\mathrm{T_{lost}}$ even though it does not make a single correct prediction of the target position. 
The basic AUC, on the other hand, correctly assigns a score 0 to this tracker.
 
Like the AUC$_\mathrm{mod}$, the proposed tracking F-measure is capable of distinguishing between $\mathrm{T_{gt,gt}}$ and $\mathrm{T_{gt,co}}$. 
In contrast to AUC$_\mathrm{mod}$, the basic primary measures, tracking Precision/Recall, offer a clear interpretation of the reason for the performance difference. 
The high tracking Precision of $\mathrm{T_{gt,gt}}$ indicates a better target absence prediction compared to $\mathrm{T_{gt,co}}$. 
But both trackers equally accurately predict target position when visible, which results in an equally high tracking Recall. Another example is the tracker $\mathrm{T_{lost}}$. 
The tracking F-score is zero, indicating a complete tracking failure and the Recall zero means that the reason is inability to localize the target.

\subsection{Robustness to annotation sparsity}  \label{sec:sparse_analysis}

\begin{figure*}[!t]
\begin{center}
	\includegraphics[width=\linewidth]{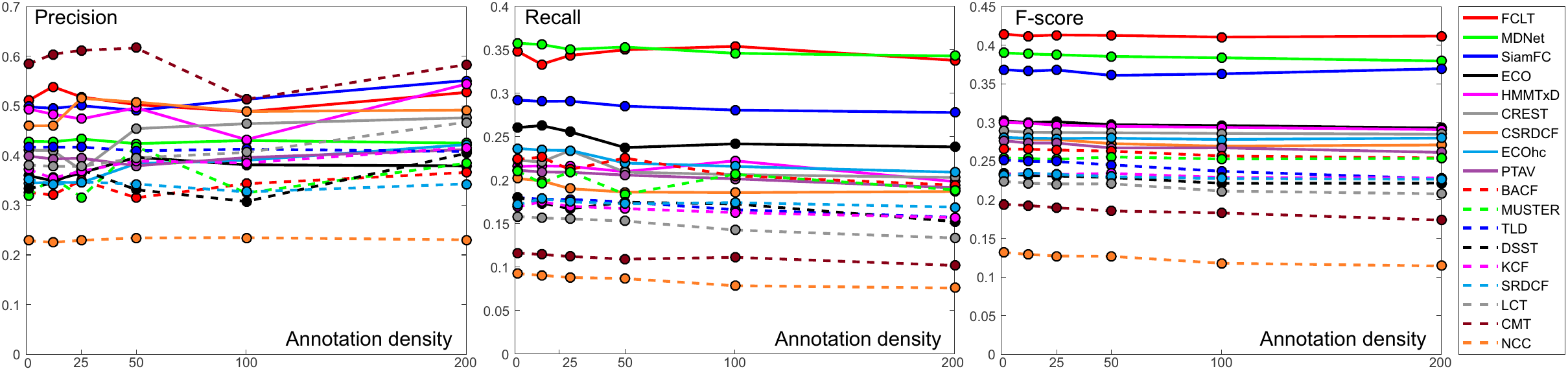}
\end{center}
   \caption{Sparsely sampled ground truth simulation by considering every $N$-th (annotation density) frame in computation of tracking precision, recall and F-score. Our primary measure, the F-score, stays extremely stable even for a very sparse annotation.} 
\label{fig:sparsity} 
\end{figure*}

Manual annotation of every frame in long-term sequences requires a significant amount of manual labor, since these are often an order of magnitude longer than short-term sequences. 
An approach to reduce the labor is annotating every $N$-th frame~\cite{Valmadre_2018_ECCV}. The amount of skipped frames is typically constrained by the robustness of the performance measure. 
We utilize densely annotated long-term sequences in LTB50 to test the behavior of the performance measures introduced in Section~\ref{sec:methodology} with respect to the annotation sparsity.

A set of trackers described in Section~\ref{sec:trackers} was run on the LTB50 dataset.
The trackers were evaluated by computing the tracking Precision, Recall and F-measure by considering every $N$-th frame with $N \in [1, 12, 25, 50, 100, 200]$. This is equal to annotating every  
0.04s, 0.5s, 1s, 2s, 4s and 8s assuming a 25fps frame rate. Figure~\ref{fig:sparsity} shows the behavior of the performance measures with increasing annotation sparsity. 

The tracking Precision/Recall deviate a bit at very high annotation sparsity levels, but largely maintain the order of trackers. A striking result is that
the deviations in Precision/Recall appear to cancel out in tracking F-score, which maintains extremely stable results over \textit{the whole range} of annotation sparsity levels. 

These results indicate that detailed performance analysis can be carried out with annotations every 25 or 50 frames since the measure values minimally differ from those obtained from dense annotations. 
This is a very important finding and means that sequences 50 times longer than typical short-term sequences can be annotated with the same amount of manual labor without losing analysis accuracy. 
If only an overall performance analysis is required (tracking F-measure) then the annotation may be even sparser, allowing \textit{up to 200 times longer sequences} at a moderate annotation effort.

\section{Long-term tracking evaluation}  \label{sec:lt-benchmark}

\subsection{Evaluated trackers}  \label{sec:trackers}

\begin{table*}[t!]
\begin{center}
\caption{\textbf{Evaluated trackers} are characterized by the short-term component and a confidence score. Long-term trackers are in addition characterized by the detector type and its interaction with the short-term component. Model update and search strategies are indicated. Trackers marked by $^*$ were published as $LT_1$, but did not pass the re-detection test. Results for the re-detection experiment are shown in the last two columns -- the number of sequences with successful re-detection (out of 50) and the average number of frames before re-detection.}
\label{tab:trackers}
\scalebox{.92}{
\begin{tabular}{l   c  c  c  c  c  c  c  c}
\rowcolor{lgray} \hline
{\bf Tracker} & {\bf S-L} & {\bf Detector} & \specialcell{{\bf Short-term} \\ {\bf component}} & \specialcell{{\bf Interaction} \\ {\bf Score}} & {\bf Update} & {\bf Search} & \specialcell{{\bf Redet.} \\ {\bf Success}} & \specialcell{{\bf Redet.} \\ {\bf Frames}} \\

\rowcolor{dgray} \hline
\specialcelll{TLD \\ \scriptsize{\cite{kalal_pami}}} & $LT_1$ & \specialcell{Random \\fern} & Flow & \specialcell{P-N learning \\ Score: conser. sim.} & \specialcell{Positive, \\negative samp.} & \specialcell{Entire image \\ (cascade)} & 18 & 0.0 \\

\rowcolor{lgray} 
\specialcelll{MUSTER \\ \scriptsize{\cite{muster_cvpr2015}}} & $LT_1$ & \specialcell{Keypoints \\ (SIFT)} & CF & \specialcell{F-B, RANSAC \\ Score: max. corr.} & \specialcell{ST: every frame \\ LT: when confident} & \specialcell{Entire image \\ (keypoint matching)} & 41 & 0.0 \\

\rowcolor{dgray} 
\specialcelll{FCLT \\ \scriptsize{\cite{fclt_arxiv}}} & $LT_1$ & CF (reg.) & CF (reg.) & \specialcell{Resp. thresh., \\Score: resp. quality} & \specialcell{ST: when confident \\ LT: mix ST + LT} & \specialcell{Entire image \\ (correlation + motion)} & 50 & 76.8 \\

\rowcolor{lgray} 
\specialcelll{CMT \\ \scriptsize{Nebehay et al~\cite{CMT_CVPR2015}}} & $LT_1$ & \specialcell{Keypoints \\(static)} & \specialcell{Keypoints \\(flow)} & \specialcell{F-B, clustering, \\correspondencies\\ Score: \# keypoints} & \specialcell{ST: always \\ LT: never} & \specialcell{Entire image \\ (keypoint matching)} & 42 & 0.0 \\

\rowcolor{dgray} 
\specialcelll{HMMTxD \\ \scriptsize{\cite{Vojir_CVIU2016}}} & $LT_1$ & \specialcell{Keypoints \\(static)} & \specialcell{Flow + CF +  \\ ASMS} & \specialcell{HMM \\ Score: \# keypoints} & \specialcell{ST: when confident \\ LT: when confident} & \specialcell{Entire image \\ (keypoint matching)} & 46 & 1.7 \\

\rowcolor{lgray} 
\specialcelll{PTAV$^*$ \\ \scriptsize{\cite{ptav_iccv2017}}} & $LT_0$ & \specialcell{Siamese \\network} & \specialcell{CF \\ (fDSST)} & \specialcell{Conf. thresh,\\ const. verif. interval \\ Score: CNN score} & \specialcell{ST: always, \\ LT: never} & \specialcell{Search window \\ (enlarged region)} & 1 & 35.0 \\

\rowcolor{dgray} 
\specialcelll{LCT$^*$ \\ \scriptsize{\cite{LCT_CVPR2015}}} & $LT_0$ & \specialcell{Random \\fern} & CF & \specialcell{k-NN, resp. thresh.\\ Score: max. corr.} & \specialcell{When \\ confident} & \specialcell{Search window \\ (enlarged region)} & 0 & - \\

\rowcolor{lgray} 
\specialcelll{SRDCF \\ \scriptsize{\cite{srdcf_iccv2015}}} & $ST_0$ & - & CF & \specialcell{- \\ Score: max. corr.} & \specialcell{Always \\(exp. forget.)} & \specialcell{Search window \\ (enlarged region)} & 0 & - \\

\rowcolor{dgray} 
\specialcelll{ECO \\ \scriptsize{\cite{DanelljanCVPR2017}}} & $ST_1$ & - & \specialcell{CF \\(deep f.)} & \specialcell{- \\ Score: max. corr.} & \specialcell{Always \\(clustering)} & \specialcell{Search window \\ (enlarged region)} & 0 & - \\

\rowcolor{lgray} 
\specialcelll{ECOhc \\ \scriptsize{\cite{DanelljanCVPR2017}}} & $ST_1$ & - & CF & \specialcell{- \\ Score: max. corr.} & \specialcell{Always \\(clustering)} & \specialcell{Search window \\ (enlarged region)} & 0 & - \\

\rowcolor{dgray} 
\specialcelll{KCF \\ \scriptsize{\cite{henriques2015tracking}}} & $ST_0$ & - & CF & \specialcell{- \\ Score: max. corr.} & \specialcell{Always \\(exp. forget.)} & \specialcell{Search window \\ (enlarged region)} & 0 & - \\

\rowcolor{lgray} 
\specialcelll{CSRDCF \\ \scriptsize{\cite{Lukezic_CVPR_2017}}} & $ST_0$ & - & CF & \specialcell{- \\ Score: max. corr.} & \specialcell{Always \\(exp. forget.)} & \specialcell{Search window \\ (enlarged region)} & 0 & - \\

\rowcolor{dgray} 
\specialcelll{BACF \\ \scriptsize{Galoogahi et al~\cite{BACF_ICCV2017}}} & $ST_0$ & - & CF & \specialcell{- \\ Score: max. corr.} & \specialcell{Always \\(exp. forget.)} & \specialcell{Search window \\ (enlarged region)} & 1 & 7.0 \\

\rowcolor{lgray} 
\specialcelll{SiamFC \\ \scriptsize{\cite{siamfc_eccv16}}} & $ST_1$ & - & CNN & \specialcell{- \\ Score: max. corr.} & Never & \specialcell{Search window \\ (enlarged region)} & 0 & - \\

\rowcolor{dgray} 
\specialcelll{MDNet \\ \scriptsize{\cite{mdnet_cvpr2016}}} & $ST_1$ & - & CNN & \specialcell{- \\ Score: CNN score} & \specialcell{When confident \\ (hard negatives)} & \specialcell{Random \\ sampling} & 0 & - \\

\rowcolor{lgray} 
\specialcelll{CREST \\ \scriptsize{\cite{crest_ICCV17}}} & $ST_0$ & - & CNN & \specialcell{- \\ Score: max. corr.} & \specialcell{Always \\ (backprop)} & \specialcell{Search window \\ (enlarged region)} & 0 & - \\

\rowcolor{dgray} 
\specialcelll{DSST \\ \scriptsize{\cite{danelljan_dsst_pami}}} & $ST_0$ & - & CF & \specialcell{- \\ Score: max. corr.} & \specialcell{Always \\(exp. forget.)} & \specialcell{Search window \\ (enlarged region)} & 0 & - \\

\rowcolor{lgray} 
\specialcelll{NCC} & $ST_1$ & - & Correlation & \specialcell{- \\ Score: max. corr.} & Never & \specialcell{Search window \\ (enlarged region)} & 0 & - \\

\rowcolor{dgray} \hline
\end{tabular}
}
\end{center}
\end{table*}

An extensive collection including top-performing trackers was complied to cover the short-term--long-term spectrum. In total, eighteen trackers summarized in Table~\ref{tab:trackers} and Figure~\ref{fig:circle_plot} were evaluated. 
We included seven long-term state-of-the-art trackers with publicly available source code: 
(i) TLD \cite{kalal_pami}, which uses optical flow for short-term component and normalized-cross-correlation for detector and a P-N learning framework for detector update. 
(ii) LCT~\cite{LCT_CVPR2015} and 
(iii) MUSTER \cite{muster_cvpr2015} that
use a discriminative correlation filter for the short-term component and random ferns and keypoints, respectively, for the detector. 
(iv) PTAV \cite{ptav_iccv2017}, that uses a correlation filter for the short-term component and a CNN retrieval system \cite{sint_cvpr16} for the detector. 
(v) FCLT \cite{fclt_arxiv}, that uses a correlation filter for both, the short-term component and the detector. 
(vi) CMT \cite{CMT_CVPR2015}, that uses optical flow for the short-term component and key-points for the detector. 
(vii) HMMTxD \cite{Vojir_CVIU2016}, that applies an ensemble of short-term trackers and a keypoint-based detector. 
These trackers further vary in the frequency and approach for model updates (see Table~\ref{tab:trackers}).

In addition to the selected long-term trackers, we have included a baseline NCC tracker \cite{kristan_vot2013} and recent state-of-the art short-term trackers: 
the standard discriminative correlation filters KCF \cite{henriques2015tracking} and DSST \cite{danelljan_dsst_pami}, 
four recent advanced versions SRDCF \cite{srdcf_iccv2015}, CSRDCF \cite{Lukezic_CVPR_2017}, BACF \cite{BACF_ICCV2017}, ECOhc \cite{DanelljanCVPR2017} 
and the top-performer on the OTB \cite{otb_cvpr2013} benchmark ECO \cite{DanelljanCVPR2017}. 
Two state-of-the-art CNN-based top-performers from the VOT \cite{kristan_vot2016} benchmark SiamFC \cite{siamfc_eccv16} and MDNet \cite{mdnet_cvpr2016} and a state-of-the-art CNN-based tracker CREST \cite{crest_ICCV17} were included as well. 
All these short-term trackers were modified to be LT$_0$ compliant, i.e., able to report the target absence. A reasonable score was identified in each tracker and used as the target prediction certainty score to detect tracking failure. 
All trackers were integrated in the VOT \cite{kristan_vot_tpami2016} toolkit for automatic evaluation.

\subsection{Re-detection experiment}  \label{sec:redetection-experiment}

\begin{figure}[!t]
\begin{center}
	\includegraphics[width=\linewidth]{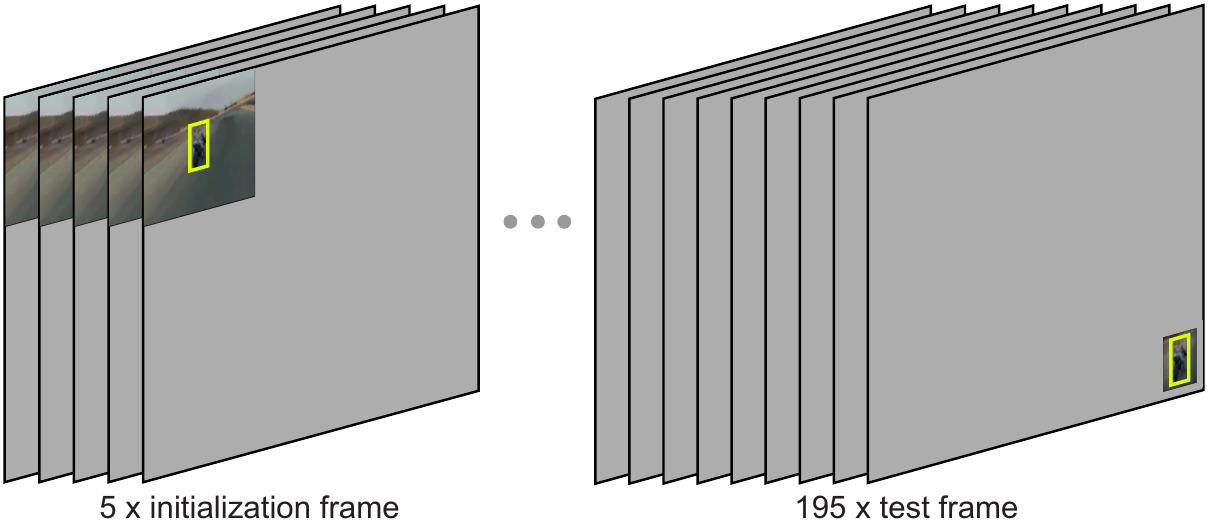}
\end{center}
   \caption{
   Re-detection experiment -- the artificially created sequence structure by repetition, padding and target displacement. Please see the text for experiment details.
   } 
\label{fig:redetection-image}
\end{figure}

An experiment was designed to position the tested trackers on the LT/ST spectrum, and in particular to verify their image-wide re-detection capability. Artificial sequences were generated from the initial frame of each sequence in our dataset.
The initial frame was placed into the top-left corner of a zero-initialized image, which is three times wider and higher than the original image (Figure~\ref{fig:redetection-image}).
The artificial sequences start with five copies of the enlarged frame. 
For the remainder of the sequence, the target region was cropped from the initial image and copied to the bottom right corner of a zero-initialized frame. 
A tracker was initialized in the first frame and we measured the number of frames required to re-detect the target after position change. 

Results are summarized in Table~\ref{tab:trackers} (last two columns). 
Trackers MDNet, ECO, ECOhc, SRDCF, SiamFC, CREST, CSRDCF, KCF, DSST and NCC never re-detected the target, which confirms their short-term design. 
The BACF tracker re-detects the target in one sequence by coincidence (random drift) and it is not the result of a re-detection mechanism.
The only tracker that always  re-detected the target was FCLT, while HMMTxD, MUSTER, CMT and TLD were successful in most sequences -- this result classifies them as $LT_1$ trackers. 
The difference in detection success come from the different detector design. FCLT and TLD both train template-based detectors. 
The  good performance of FCLT likely comes from the efficient discriminative filter training framework of the FCLT detector. 
The keypoint-based detectors in HMMTxD, MUSTER and CMT are similarly efficient, but require sufficiently well textured targets. Interestingly, the re-detection is immediate for MUSTER, CMT, HMMTxD and TLD, while FCLT requires on average 77 frames. 
This difference comes from the dynamic models. Muster, CMT, HMMTxD and TLD apply a uniform spatial prior in the dynamic model in the detector phase over the entire image, while the FCLT applies a random walk model that gradually increases the target search range with time.

Surprisingly, two recent long-term trackers, LCT and PTAV nearly never successfully detected the target. 
A detailed inspection of their source code revealed that these trackers do not apply their detector to the whole image, but rather a small neighborhood of the previous target position, which makes these two trackers a pseudo long-term, i.e., $LT_0$ level. 

\subsection{Overall performance} \label{sec:overall_evaluation}

\begin{figure}[!t]
\begin{center}
	\includegraphics[width=1\linewidth]{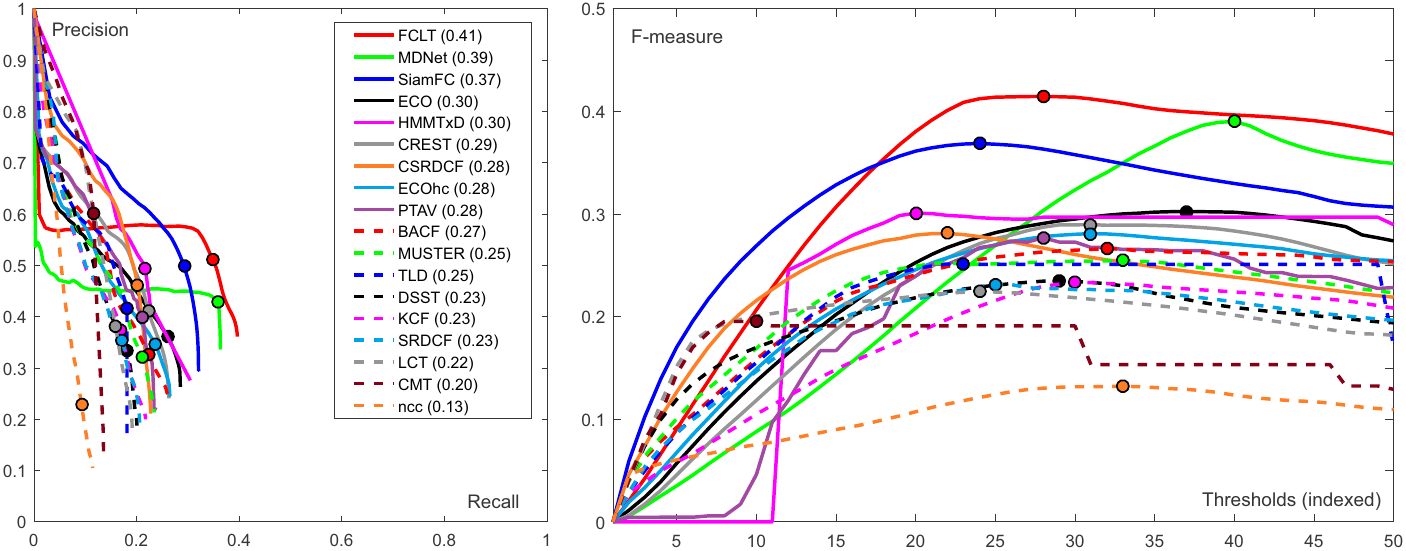}
\end{center}
   \caption{Long-term tracking performance on the LTB50 dataset. The average tracking precision-recall curves (left), the corresponding F-measure curves (right) -- F as a function of prediction certainty linearly rescaled for each tracker; 0 - the minimum over all sequences output by a given tracker, 100 - the maximum. Tracker labels are sorted according to F-scores, i.e., F-measure maxima. } 
\label{fig:average_f_pr_re} 
\end{figure}

The overall performance on the LTB50 dataset is summarized in Figure~\ref{fig:average_f_pr_re}. The highest ranked is FCLT, an LT$_1$ class tracker, which uses discriminative correlation filters on hand-crafted features for short-term component as well as detector in the entire image. 
Interestingly FCLT is followed by three \textit{short-term} ST$_1$ class CNN-based trackers MDNet, SiamFC and ECO. 
These implement different mechanisms to deal with occlusion. MDNet applies very conservative updates, SiamFC does not update the model at all and ECO applies clustering-based update mechanism to prevent learning from outliers. 
SiamFC, ECO and MDNet search a fairly large region which is beneficial for target re-detection.

Another LT$_{1}$ long-term tracker, HMMTxD, achieves comparable performance to ECO. It uses an ensemble of short-term trackers with weak visual models, and performs image-wide target re-detection.
Two long-term trackers CMT (LT$_1$) and LCT (LT$_0$) perform the worst among the tested trackers. The CMT entirely relies on keypoints, which perform poorly on non-textured targets. 
The relatively poor performance of LCT is likely due to a small search window and poor detector learning. 
This is supported by the fact that LCT performance is comparable to KCF, a standard correlation filter, also used as the short-term component in LCT. The performance of short-term trackers ST$_0$ class trackers does not vary significantly.

\subsection{Attribute evaluation}  \label{sec:attribute_evaluation}

\begin{figure}[t]
\begin{center}
	\includegraphics[width=1\linewidth]{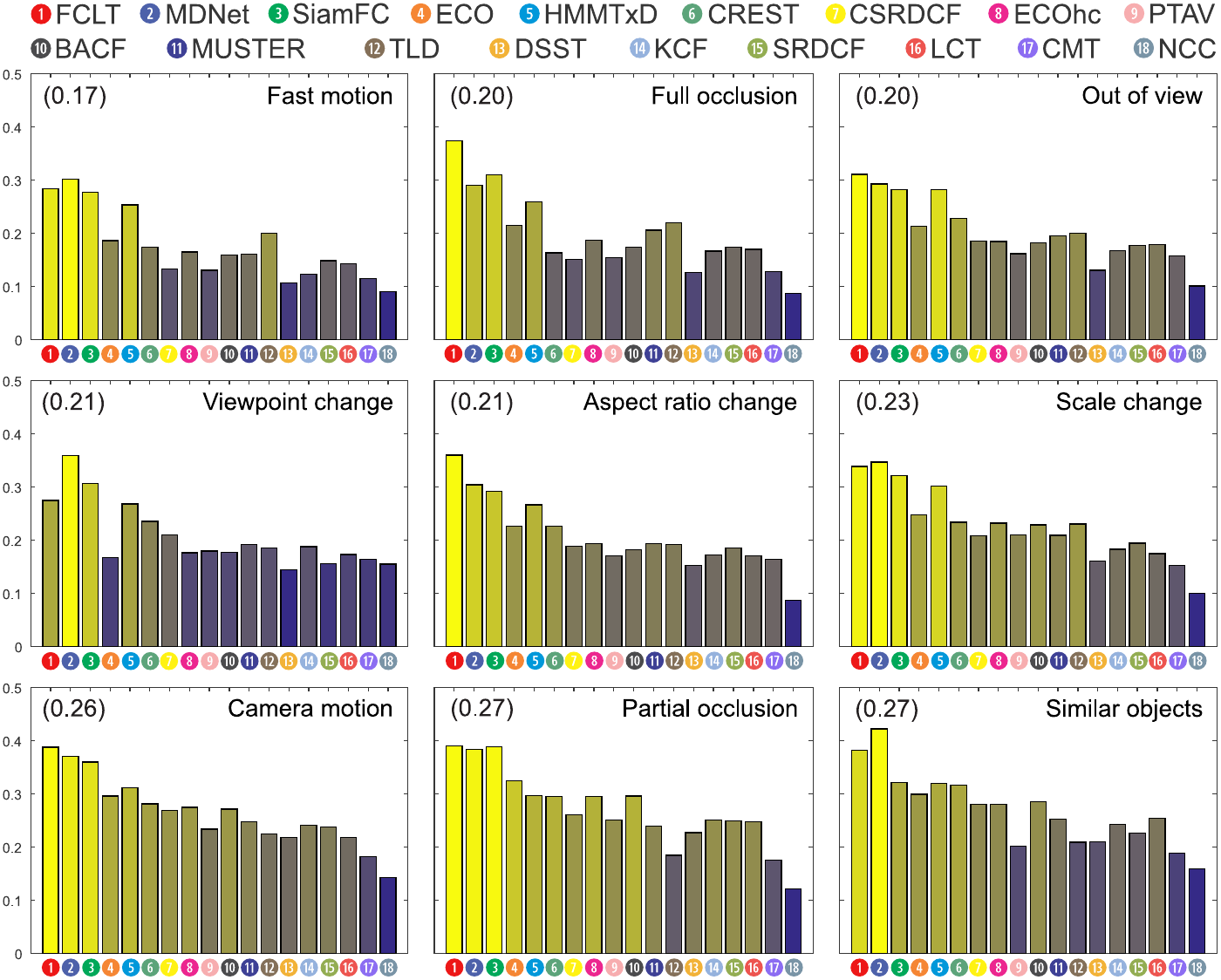}
\end{center}
   \caption{Average per-tracker F-scores for each visual attribute. Average F-score of each attribute is shown at the top-left corner. The most challenging attributes are fast motion, full occlusion and out-of-view.} 
\label{fig:attributes_average_f} 
\end{figure}

Figure~\ref{fig:attributes_average_f} shows tracking performance with respect to nine visual attributes from Section~\ref{sec:dataset}. 
Long-term tracking is mostly characterized by performance on {\it full occlusion} and {\it out-of-view} attributes, since these require re-detection. 
The FCLT (LT$_1$ class) achieves top performance, which is likely due to the efficient learning of the detector component. 
Another LT$_1$ tracker, HMMTxD, performs comparably to the best short-term trackers (SiamFC and MDNet), while the CMT, TLD and MUSTER performance is lower due to a poor visual model.

A highly challenging attribute is {\it fast motion} which is related to long-term re-detection combined with blurring. Top performance is obtained by trackers with a relatively large search range (FCLT, SiamFC, MDNet).
 
Another attribute specific for long-term tracking is {\it viewpoint change} which includes video cuts and camera hand-overs. 
Top-performing trackers at this attribute are SiamFC and MDNet, which indicates that in most of these viewpoint changes the target did not move significantly in image coordinates, and moderate search range sufficed in target re-detection. The result also shows that the target appearance did change and was well addressed by the deep features.

The {\it similar objects} attribute exposes fine-grained discrimination capability between the tracked object and visually similar objects in the vicinity. 
The top-performing tracker at this attribute is MDNet, which is likely due to use of hard-negative mining in the visual model update and a moderately sized search range.
Another well performing tracker is FCLT. In contrast to MDNet, this tracker performs image-wide re-detection, which increases the probability of drifting to another object, even if the object is far away. These false detections are mitigated by the motion model, that gradually increases the effective detection range after localization becomes uncertain.

\subsection{Influence of disappearance frequency}  \label{sec:sequence_evaluation}

\begin{figure*}[t]
\begin{center}
	\includegraphics[width=1\linewidth]{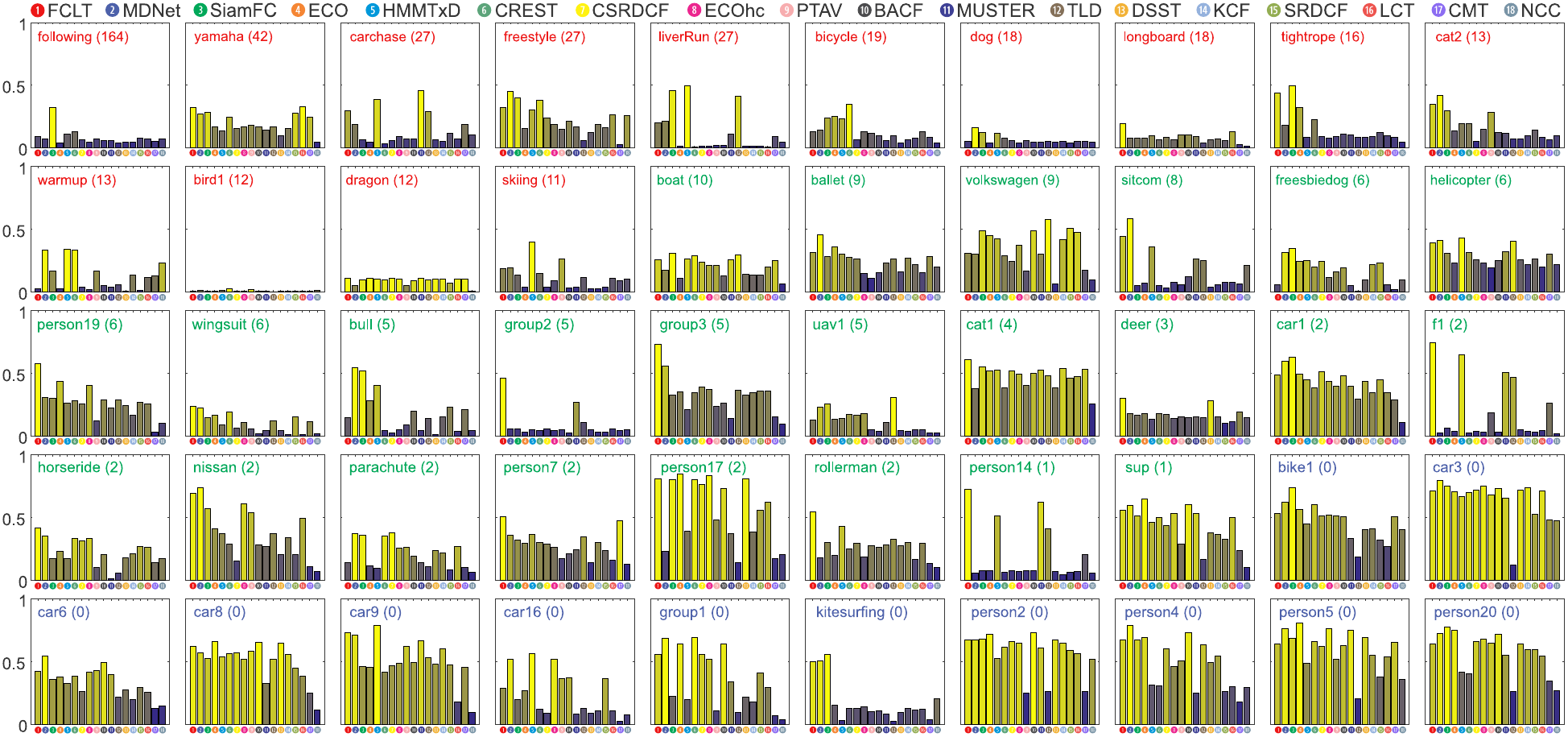}
\end{center}
   \caption{Target disappearance analysis: The plots show per-sequence F-scores for all trackers. Sequences are sorted, left-to-right, top-to-bottom, by the  number of target disappearances, i.e. the largest number at top-left. Red label:  $(>10)$ disappearances, green: $(1-10)$ disappearances, blue: (0) disappearances.} 
\label{fig:per_seq_average_f} 
\end{figure*}

We divided the sequences of LTB50 into groups according to the number of target disappearances: (Group~1) over ten disappearances, (Group~2) between one and ten disappearances and (Group~3) no disappearances. Per-sequence F-scores are summarized in Figure~\ref{fig:per_seq_average_f}. 
 
\textit{Group 1 results:} Most short-term trackers performed poorly due to lack of target re-detection. Long-term trackers generally perform well, but there are differences depending on their structure. For example, the ``following'' and ``liverrun'' sequences contain cars, which only moderately change the appearance. SiamFC does not adapt the visual model and is highly successful on these sequences. The LCT generally performs poorly, except for the ``yamaha'' sequence in which the target leaves and re-enters the view at the same location. Thus the poor performance of LCT is due to a fairly small re-detection range. MDNet, CREST and SiamFC perform moderately well, despite the fact that they are short-term trackers. A likely reason is their highly discriminative visual features (CNNs) and a relatively large target localization range.

\textit{Group 2 results:} Performance variation comes from a mix of target disappearance and other visual attributes. However, in ``person14'' the poor performance is related to a long-lasting occlusion at the beginning, where most trackers fail. Only some of LT$_1$ class trackers (FCLT, MUSTER, HMMTxD and TLD) overcome the occlusion and obtain excellent performance.
 
\textit{Group 3 results:} The performance of long-term trackers does not significantly differ from short-term trackers since the target is always visible. The strength of the features and learning in visual models play a major role. These sequences are least challenging for all trackers in our benchmark.

\subsection{Tracking speed analysis}  \label{sec:speed-analysis}

Tracking speed is a decisive factor in many applications. We provide a detailed
analysis by three measures\footnote{Due to the limitations of the source code of MUSTER provided by the authors we were able to calculate the average speed, but not initialization and maximum per-frame times.}: 
(i) initialization time, (ii) maximum per-frame time and (iii) average per-frame time. 
The initialization time is computed as the initial frame processing time averaged over all sequences. 
The maximum per-frame time is computed as the median of the slowest 10\% of the frames averaged over all sequences.
We also measure average speed by averaging over all frames in the dataset. All measurements are in milliseconds per frame (mpf). 
The experiments were carried out at a standard desktop computer with 3.4GHz 6700-i7 CPU, 16GB of RAM and NVidia GTX 1060 GPU with 6GB of RAM.

\begin{figure}[t]
\begin{center}
	\includegraphics[width=\linewidth]{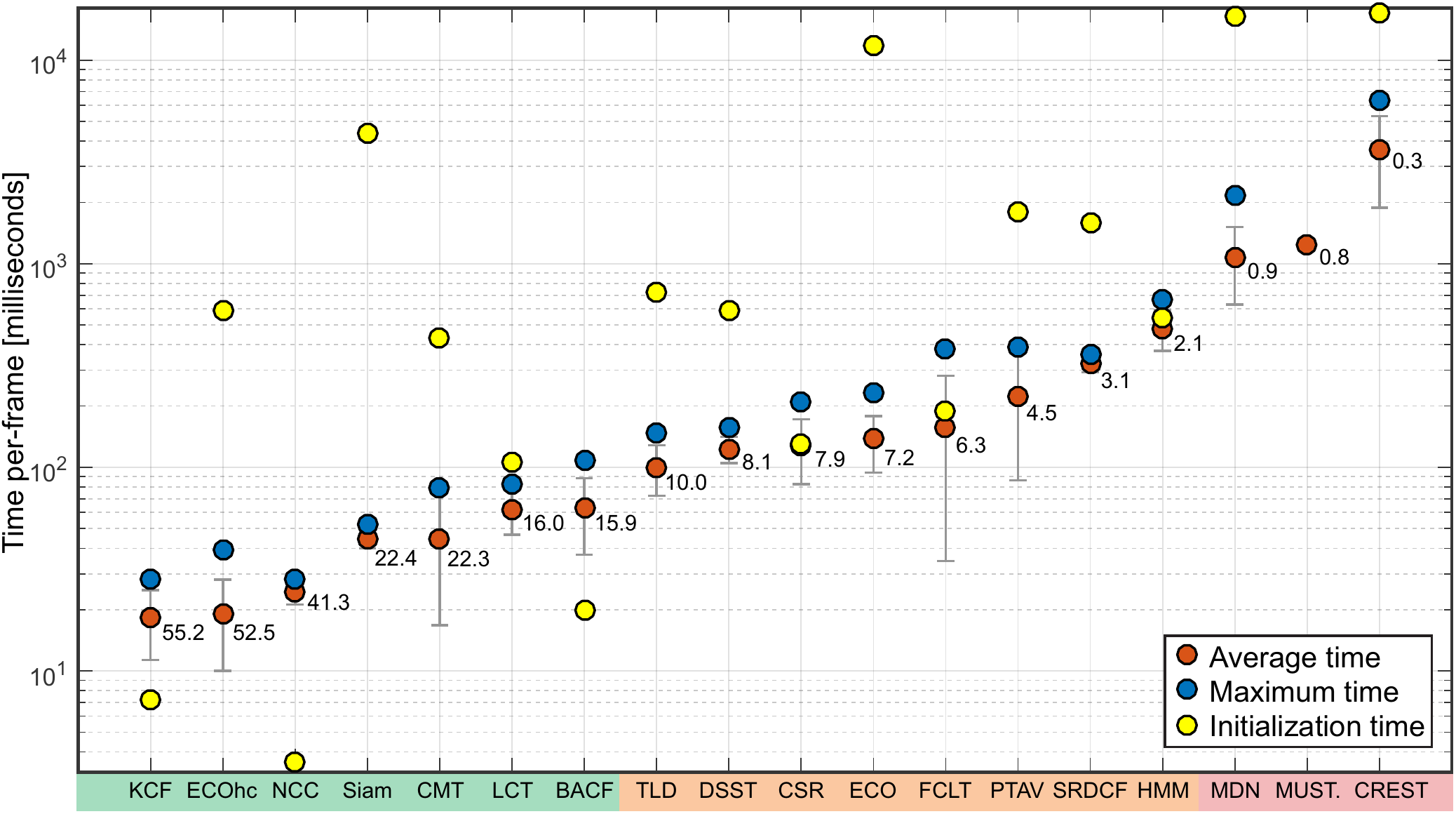}
\end{center}
   \caption{Tracking speed. Trackers are grouped into three classes: fast (green), moderately fast (orange) and slow (red). All measurements are in milliseconds-per-frame (mpf) -- initialization (yellow), average (red), maximum (blue). Average speed measured as frames-per-second is shown next to the position of average mpf. Note that logarithmic scale is used on the y-axis.} 
\label{fig:time_analysis} 
\end{figure}

The tracking speeds are reported in Figure~\ref{fig:time_analysis} with trackers categorized into three groups according to the average speed: fast ($>15$fps), moderately fast (1fps-15fps) and slow ($<1$fps). The fastest tracker is KCF due to efficient model learning and localization by fast Fourier transform. 
The slowest methods are CNN-based MDNet and CREST due to the time-consuming model adaptation and MUSTER due to slow keypoint extraction in the detection phase. 
Several trackers exhibit a very high initialization time (in order of several thousand mpf). The delay comes from loading CNNs (SiamFC, ECO, PTAV, MDNet, CREST) or pre-calculating visual models (ECOhc, CMT, TLD, SRDCF, DSST).

Ideally, the tracking speed would be approximately constant over all frames guaranteeing completion within a fixed time delay. Small differences between the maximum per-frame and average time indicate stability. 
This difference is the largest for the following trackers: ECOhc and ECO (due to a time-consuming update every five frames), 
FCLT (due to re-detection on the entire image, which is moderately slow for large images), 
PTAV (due to the slow CNN-based detector) and MDNet (due to the slow update during reliable tracking period).

\section{Tracker architecture evaluation}  \label{sec:architecture_evaluation}

\subsection{Overall architecture analysis}

\begin{figure}[htb]
\begin{center}
	\includegraphics[width=\linewidth]{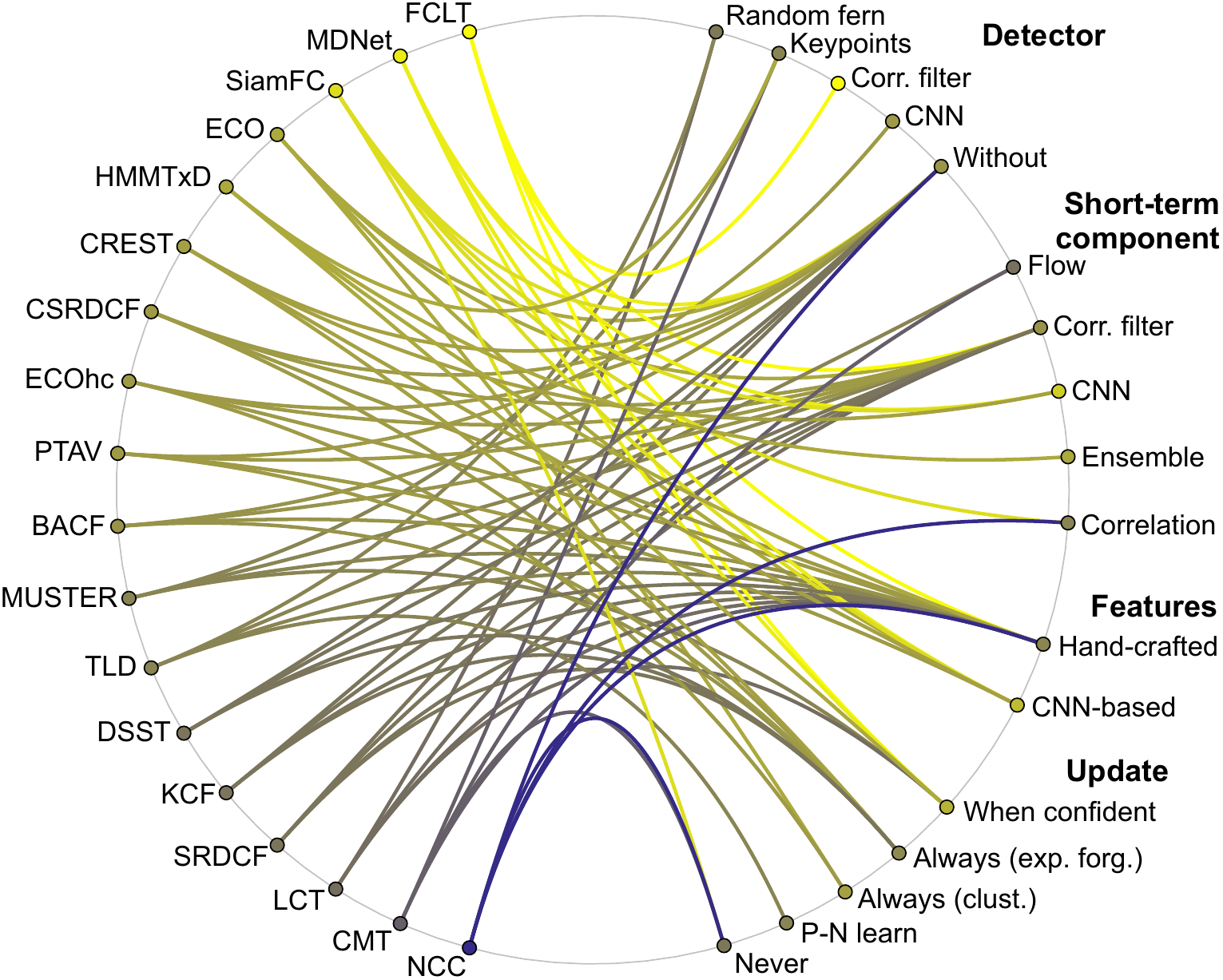}
\end{center} 
   \caption{The connection plot visualizing the tracker architectural choices. The trackers are linked to the particular implementations of the long-term tracking components. The links are color coded by the tracking F-score on LTB50 dataset with yellow indicating high values blue indicting low values.} 
\label{fig:circle_plot} 
\end{figure}

We analyze contributions of architectural choices important for successful long-term tracking by categorizing the tested trackers along the following four aspects: (i) detector design, (ii) short-term component design, (iii) features used and (iv) visual model adaptation strategy. To aid interpretation, we generate a connection plot 
Figure~\ref{fig:circle_plot} where each tracker is connected to the specific choice of the four design aspects thus visualizing a design trend by color-coding. 

\textbf{Detector design:} The results show that CNN-based detectors consistently deliver promising performance. 
Correlation filters are widely used in short-term trackers, but are generally not used for image-wide detection, except for in the FCLT. The plot indicates that the kind of DCF-based detector used in FCLT might be a very promising research direction in long-term tracker design. 
The quality of keypoint-based detectors varies significantly among the trackers. The benefit lies in potential to re-detect target even under a similarity or affine transform, but a common drawback is the inability to detect small or homogeneous targets.

\textbf{Short-term component:} The most promising design choices for the short-term component follow the trend in state-of-the-art short-term tackers. The connection plot indicates that CNN-based and DCF-based methods are most successful short-term design choices.  

\textbf{Visual features:} Visual models with features based on CNN generally achieve improved performance over hand-crafted features. The reason is likely in discriminative capacity of the pre-trained networks. 
A drawback is that these features typically entail significant computational resources and dedicated hardware (i.e., GPU). Results also show that CNN features are not crucial for high-quality long-term tracking. 
In fact hand-crafted features combined with a well-designed re-detection strategy or update mechanism (e.g., FCLT) on average by far outperform all CNN-based trackers.

\textbf{Adaptation strategy:} In long-term tracking scenarios, the target may leave the field of view or become occluded for longer periods. 
Constant updating irreversibly corrupts the visual model leading to drift and failure and reduces the potential for target re-detection. 
Conservative updating such as implemented in MDNet or FCLT appears to be the best strategy. An extreme conservative update strategy, i.e., no update at all, appears to work as well, but this requires highly expressive features such as localization-trained CNN in SiamFC.

\subsection{Importance of re-detection strategy}  \label{sec:import_redet_strategy}


We further explore the importance of re-detection strategy by the following experiment. All tracker outputs were modified to report a constant uncertainty, which was treated as if the tracker is always reporting the target as present. Tracking Recall was computed, which we denote by $Re$. Then for each tracker, all overlaps after the first failure (i.e., overlap drops to zero) were set to zero and the tracking Recall was re-computed ($Re_0$).

\begin{figure}[t]
\begin{center}
	\includegraphics[width=\linewidth]{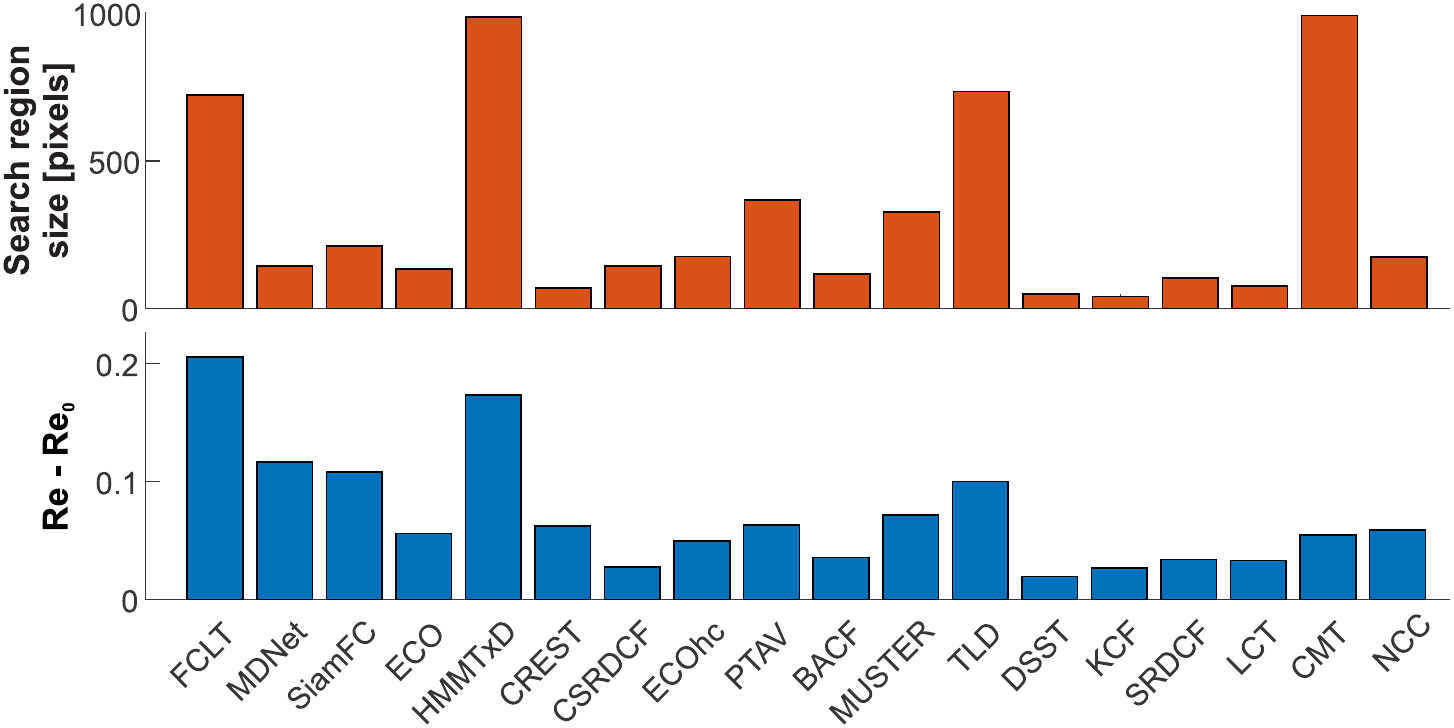}
\end{center} 
   \caption{The red graph shows results of the re-detection search range size experiment (above). Higher values indicate a larger search range. The blue graph (below) shows the difference in tracking Recall between the original tracking result and the one with all overlaps set to zero after the first failure. Large values indicate increased influence of the re-detection.} 
\label{fig:redetection-live} 
\end{figure}

Figure~\ref{fig:redetection-live} shows differences between the two recalls (i.e., $Re - Re_0$). Large values indicate greater failure recovery capabilities of the trackers. FCLT, HMMTxD and TLD most often re-detected the target. 
Surprisingly, the differences in tracking Recalls of two short-term trackers MDNet and SiamFC are very large, which indicates that these two trackers indeed posses long-term properties. There are two probable explanations: (i) the trackers posses a large search region which enables target re-detection or (ii) the trackers posses efficient visual model update mechanism
that prevents visual model corruption during target loss and they eventually drift back to the target. 
 
Additional analysis of the tracking performance was carried out to determine the reason for the apparent long-term properties. Let $o_{i}$ denote an overlap at $i$-th frame of a sequence. We identified the pair of frames where $o_{i-1} = 0$ and $o_{i} > 0$ which is a point at which the tracker re-detects the target. 
Euclidean distance was computed between the predicted bounding box centers in these frames. We expect that large distances indicate large target search ranges and compute the mean value of the ten percent of largest Euclidean distances as an indicator of the recorded search range size.

The results are shown in Figure~\ref{fig:redetection-live}. The search range size of MDNet and SiamFC are comparable to other short-term trackers.
This means that the key factor for their excellent long-term tracking performance compared to the other short-term trackers is the visual model. Both of these trackers use pre-trained CNN-based features. 
MDNet updates the visual model only on frames where tracking is considered reliable while SiamFC does not update the visual model at all. 
Both mechanisms prevent training from incorrect examples, which enables eventual re-detection once the target gets close to the current tracker prediction, even though the search range is not the whole image.
Long-term trackers FCLT, TLD, CMT and HMMTxD have the largest search range, which confirms the image-wide target re-detection ability tested in Section~\ref{sec:redetection-experiment}. 
MUSTER and PTAV have a moderately large search region, while LCT has the smallest search range among all long-term trackers.

\subsection{Impact of visual model error accumulation}  \label{sec:import_update_strategy}

Visual model update strategy plays a central role in drift prevention, which is crucial in tracking over long periods. We designed the following experiment to evaluate drifting, which is not caused by target disappearance. 
Based on the disappearance frequency analysis from Section~\ref{sec:sequence_evaluation}, we selected eleven long sequences from LTB50 in which the target never disappears and extended them by looping forward and backward five times. 
This set of extended sequences thus contains 302,330 frames with an average sequence length of 27,485 frames.   

The trackers were re-run on this dataset and the tracking Recall was computed with the same tracker output modification to a constant value as in Section~\ref{sec:import_redet_strategy}. 
Since the target never leaves the field of view and the tracker always reports the target, the tracking Recall is equivalent in this case to the average overlap on all frames (see Section~\ref{sec:methodology}).
 
\begin{figure}[t]
\begin{center}
	\includegraphics[width=\linewidth]{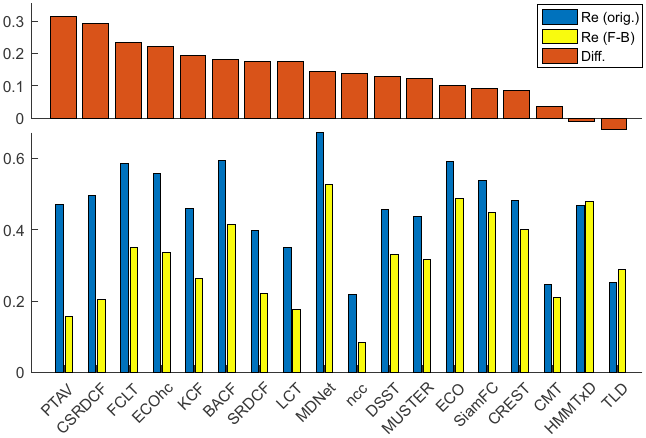}
\end{center} 
   \caption{The bottom plot shows tracking recall of trackers evaluated on sequences from LTB50 with target always visible (blue) and the same measure when running the trackers on the same sequences forward/backward five times (yellow). The differences between these recalls indicate tracker sensitivity to long sequences due to error accumulation in the visual model and are shown in the upper plot (red). Larger difference represents larger impact of the sequence length. Note that the error increases for all trackers except from TLD, that learns from experience and actually decreases the error.} 
\label{fig:f-b} 
\end{figure}

The results are shown in Figure~\ref{fig:f-b}. The highest tracking recall is achieved by two ST$_1$ short-term trackers MDNet, ECO, an ST$_0$ tracker BACF and an LT$_1$ long-term tracker FCLT. 
Top positions are dominated by state-of-the-art short-term trackers in part because the target is always present and false activations of the detector during uncertain target localization periods may lead to tracker jumping to a location away from the target, which reduces performance.

Performance drop is smallest for TLD, HMMTxD, CMT, CREST and SiamFC. 
The CMT fails early on in the original sequences, which explains the apparently small performance drop. CREST applies end-to-end updating of all parameters in a CNN with small learning rate, which leads to robust tracking in situations when the target is always visible. 
The small performance drop in SiamFC is likely due to the fact that this tracker does not update the visual model. Combined with the deep features, this proves as a robust strategy to reduce visual model contamination and leads to a successful tracking. 
Interestingly, the recall actually increases for TLD and HMMTxD as the sequences are looped, which is consistent with the observations in the original paper~\cite{kalal_pami}. 
TLD applies P-N learning, a conservative form of learning that retrospectively expands the visual model with new training examples. The longer the target is observed, the stronger the visual model becomes. Similarly, the HMMTxD uses combination of feature-based detector, which is trained only in the initial frame and set to high precision mode, that guides the on-line learning of the hidden Markov model. The HMM encodes the relationship of the performance of individual trackers and their confidences using Baum-Welch algorithm. This combination enables choosing, on-the-fly, which tracker should be used in every frame and improves over time.
The selective update strategies from MDNet and LCT, which mix short-term and long-term updates also appear beneficial -- MDNet, for example, actually keeps track of appearance samples from a longer time-scale and uses these in combination with a local hard negative mining in the model update. 

The largest performance drop is observed for the long-term trackers PTAV and FCLT and the short-term trackers CSRDCF and ECOhc. 
There are several reasons for these performance drops. 
All four trackers use a DSST~\cite{danelljan_dsst_pami} scale estimation method. We observed that the scale at these trackers gradually drifts in extremely long sequences. 
The reason might be that DSST scale estimation relies very much on the target localization accuracy. 
Inacurate localization leads to incorrect scale estimation, and gradual error accumulation from constant scale updates further reduces the localization accuracy, leading to drift.

The long-term FCLT and PTAV are affected by false activations of their detector, which in some cases leads to tracker jumping off the target. 
FCLT updates the detectors over several scales only during certain tracking periods and eventually re-detects the target in most cases. PTAV applies a CNN instance-based object detector without updates. 
The strength of this detector is that it generalizes well enough to detect the target even under deformation. On the other hand, this generalization leads to failure when similar objects are located in the target vicinity. 
This is an obvious reason in the \textit{bike1} sequence where the detector jumps to another bicyclist.

There is a significant difference in performance drops of ECO and ECOhc, even though they both use the same visual model decontamination strategy during updates. 
Part of the difference can be attributed to different scale estimation strategies these two trackers use. ECO applies the tracker over several scales, while ECOhc applies the DSST on the estimated target position. 
Another significant difference is that ECO applies deep features, while ECOhc uses only HOG and Colornames. Therefore the longer tracking periods observed in ECO might be likely due to deep features and greedy scale search.

\section{Discussion and conclusion}  \label{sec:conclusion}

A new long-term single-object tracking benchmark was presented. 
A new short-term/long-term tracking taxonomy that predicts performance on sequences with long-term properties was introduced. The taxonomy considers (i) the target absence prediction capability, (ii) the target re-detection strategy and (iii) the visual model update mechanism.
A new long-term tracking performance evaluation methodology which introduces new performance measures -- {\em tracking} Precision, Recall and F-score -- is proposed. 
These measures extend the detection analysis capabilities to tracking in a principled way and theoretically link the short-term and long-term tracking problem.
A new dataset (LTB50) of carefully selected sequences is constructed, with a significant number of target disappearances per sequence to emphasize long-term tracking properties. 
Our experiments in Section~\ref{sec:sequence_evaluation} indicate that target disappearance is in fact the most challenging aspect of long-term tracking. The diversity of the dataset has been ensured by including a number of target examples typical for long-term tracking in a variety of environments. 
Sequences are annotated with nine visual attributes which enable in-depth analysis of trackers. 
Seven long-term trackers and eleven state-of-the-art short-term trackers were categorized using the new taxonomy and analyzed using the proposed methodology and the dataset.
 
Comparison with existing performance measures using theoretical trackers (Section~\ref{sec:measures-comparison}) shows that the proposed tracking Precision, Recall and F-score outperform existing measures, distinguish well between different tracking behaviors and facilitate its interpretation. 
Furthermore, these measures are highly robust (Section~\ref{sec:sparse_analysis}) allowing detailed analysis with only every 50th frame annotated. 
The overall ranking based on the primary measure is even more robust allowing even sparser annotations (e.g., every 200th frame). 
This is a significant result since it shows that datasets with up to 200 times longer sequences can be annotated with equal manual annotation effort without interpolation.

The evaluation and analysis covers a comprehensive collection of long-term trackers. According to the overall analysis (Section~\ref{sec:overall_evaluation}), the best performance is obtained by a LT$_1$ long-term tracker FCLT~\cite{fclt_arxiv}. 
This tracker applies discriminative correlation filter  as the short-term component as well as for detector. It applies updating of the visual models at various temporal scales and uses the correlation output for predicting target absence. 
The second-best tracker is, surprisingly, a state-of-the-art short-term tracker MDNet~\cite{mdnet_cvpr2016}, which is a CNN-based tracker trained for the tracking task. It applies hard-negative mining and conservative updating of a few top-layer CNN features. 
Attribute analysis (Section~\ref{sec:attribute_evaluation}) indicates that full occlusions and out-of-view disappearances are among the most challenging attributes, followed by similar objects and viewpoint change. 
The analysis also shows that the LTB50 dataset is challenging, the best tracker achieves the average F-score of 0.41, leaving room for improvement. 

Further insights are obtained by analyzing architecture designs of the long-term trackers (Section~\ref{sec:architecture_evaluation}). 
CNN-based detectors consistently deliver improved performance, which is likely due to their expressive power of robustly localizing the target even under moderate appearance changes. 
However, appearance generalization may come at a cost when visually similar objects are located in the same scene. In these cases the CNN features may not distinguish between the different objects, leading to tracking the wrong target (Section~\ref{sec:import_update_strategy}). 
Even though discriminative correlation filters are not widely used for detectors, results show that careful learning e.g., \cite{fclt_arxiv} makes them an excellent choice due to speed and robustness. We expect to see many long-term trackers adapt these in future. 
Keypoint-based detectors can potentially detect the target under similarity transform e.g., \cite{muster_cvpr2015,Vojir_CVIU2016}, but require textured targets and sufficient resolution, which makes them brittle in practice.

The re-detection experiments from Section~\ref{sec:redetection-experiment} and Section~\ref{sec:import_redet_strategy} show that most successful re-detection strategies are those used in FCLT \cite{fclt_arxiv}, HMMTxD \cite{Vojir_CVIU2016} and TLD \cite{kalal_pami}. 
Results in Section~\ref{sec:import_redet_strategy} also show that re-detection quality largely depends on the visual model update strategy. 
Conservative updates \cite{siamfc_eccv16} and hard-negative mining \cite{mdnet_cvpr2016} show promise. These techniques are crucial for tracking on very long sequences even if the target is always visible (Section~\ref{sec:import_update_strategy}), since they largely reduce the tracking drift. 
This finding opens an opportunity for improving long-term tracking by considering best practices in visual features and model updating from short-term trackers. 

Scale estimation methods play an important role in tracking drift. A popular approach is to first localize the target and then estimate the scale e.g., by \cite{danelljan_dsst_pami}, 
considering only a single position. 
Trackers with this technique typically fare much worse than those that greedily localize the target on several scales. 
The reason is that inaccurate localization leads to poor scale estimation, which consequently leads to poorer localization. On long sequences, the errors accumulate in the visual model, resulting in drift.
 
Tracking speed analysis (Section~\ref{sec:speed-analysis}) shows that reporting solely the  average speed may be misleading and insufficient for applications that require short response times. 
Many trackers, especially long-term, perform very expensive re-detection or learning operations at regular or even unpredictable time instances.
Furthermore, initialization times for several trackers are order of magnitude larger than the standard tracking iteration. We recommend that additional information, like the maximum response time and initialization times should be reported as part of standard analysis. 
 
All tested trackers and performance evaluation methodology have been integrated in the VOT toolkit \cite{kristan_vot_tpami2016} and will be made publicly available to the research community. 
We believe that this, along with the evaluation methodology and detailed analysis presented in this paper, will significantly impact the field of long-term tracking from the point of dataset construction with extremely long sequences, performance analysis protocols as well as long-term tracker designs.


\bibliographystyle{IEEEtran}
\bibliography{bib}

\begin{IEEEbiography}[{\includegraphics[width=1in,height=1.25in,clip,keepaspectratio]{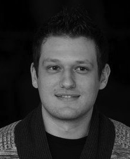}}]
{Alan Lukežič} 
received the Dipl.ing. and M.Sc. degrees at the Faculty of Computer and Information Science, University of Ljubljana, Slovenia in 2012 and 2015, respectively. 
He is currently a PhD student at Faculty of Computer and Information Science, University of Ljubljana, Slovenia. 
His research interests include computer vision, data mining and machine learning.
\end{IEEEbiography}\vspace{-1cm}

\begin{IEEEbiography}[{\includegraphics[width=1in,height=1.25in,clip,keepaspectratio]{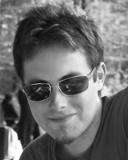}}]
{Luka Čehovin Zajc} 
received his Ph.D from the Faculty of Computer and Information Science, University of Ljubljana, Slovenia in 2015. 
Currently he is working at the Visual Cognitive Systems Laboratory, Faculty of Computer and Information Science, University of Ljubljana, Slovenia as an assistant professor and a researcher. 
His research interests include computer vision, HCI, distributed intelligence and web-mobile technologies.
\end{IEEEbiography}\vspace{-1cm}

\begin{IEEEbiography}[{\includegraphics[width=1in,height=1.25in,clip,keepaspectratio]{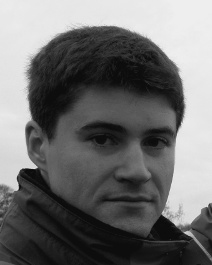}}]
{Tomáš Vojíř} 
received his Ph.D from the Faculty of Electrical Engineering, Czech Technical University in Prague, 
Czech Republic in 2018. Currently he is working at the Machine Intelligence Laboratory, Department of 
Engineering, University of Cambridge, United Kingdom as a research associate. His research interests 
include real-time visual object tracking, object detection, online learning and vision for autonomous driving.
\end{IEEEbiography}\vspace{-1cm}

\begin{IEEEbiography}[{\includegraphics[width=1in,height=1.25in,clip,keepaspectratio]{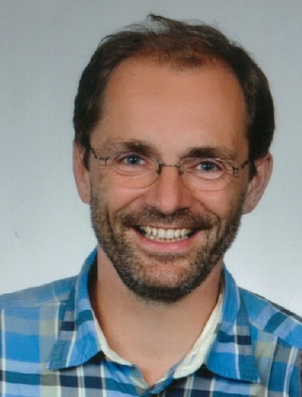}}]
{Jiří Matas} is a professor at the Center for Machine Perception,
Czech Technical University in Prague. He holds a PhD degree from the University
of Surrey, UK (1995). He has published more than 200 papers in refereed journals
and conferences. His publications have approximately 34000 citations in Google Scholar and 13000 in the Web of Science. His h-index is 65 (Google scholar) and 43 (Clarivate Analytics Web of Science)   
respectively. He received the best paper prize at the British Machine Vision
Conferences in 2002 and 2005, at the Asian Conference on Computer Vision
in 2007 and at Int. Conf. on Document analysis and Recognition in 2015.
 He is on the editorial board of IJCV and was an Associate Editor-in-Chief of IEEE T. PAMI. 
His research interests include visual tracking, object recognition,
image matching and retrieval, sequential pattern recognition, and RANSAC-
type optimization methods.
\end{IEEEbiography}\vspace{-1cm}

\begin{IEEEbiography}[{\includegraphics[width=1in,height=1.25in,clip,keepaspectratio]{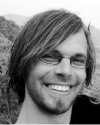}}]
{Matej Kristan} 
received a Ph.D from the Faculty of Electrical Engineering, University of Ljubljana in 2008. 
He is an Assistant Professor at the ViCoS Laboratory at the Faculty of Computer and Information Science and at the Faculty of Electrical Engineering, University of Ljubljana. 
His research interests include probabilistic methods for computer vision with focus on visual tracking, dynamic models, online learning, object detection and vision for mobile robotics.
\end{IEEEbiography}\vspace{-1cm}

\end{document}